\documentclass[10pt,twocolumn,letterpaper]{article}

\usepackage[pagenumbers]{cvpr} 

\usepackage{graphicx}
\usepackage{amsmath}
\usepackage{amssymb}
\usepackage{booktabs}

\usepackage{siunitx}
\usepackage[usenames,dvipsnames]{xcolor}
\usepackage{multicol}
\usepackage{multirow}
\usepackage{subcaption}

\usepackage{amsthm}%
\usepackage{mathrsfs}%
\usepackage[title]{appendix}%
\usepackage{textcomp}%
\usepackage{manyfoot}%
\usepackage{algorithm}%
\usepackage{algorithmicx}%
\usepackage{algpseudocode}%
\usepackage{listings}%

\usepackage{inputenc}
\usepackage{array}
\usepackage{stfloats}
\usepackage{url}
\usepackage{verbatim}
\usepackage{balance}
\usepackage{diagbox, eqparbox}
\usepackage{tabularx}
\usepackage{tabu}
\usepackage{soul}
\usepackage{makecell}
\usepackage[symbol]{footmisc}
\usepackage[normalem]{ulem}
\usepackage{threeparttable}
\usepackage{mathtools}

\usepackage{lipsum}
\usepackage{tcolorbox}
\usepackage{caption}
\usepackage{xpunctuate}

\newcommand{\Tref}[1]{Tab.~\ref{#1}}

\newcommand{\Fref}[1]{Fig.~\ref{#1}}

\newcommand{\methodname}{DisAE\xspace}
\newcommand{\mpiitarget}{MPII-NV\xspace}

\newcommand{\mvxgaze}{MVS-XGazeF-NV\xspace}

\providecommand{\etal}{\textit{et al}\xperiod}
\providecommand{\eg}{\textit{e.g}\xperiod}

\providecommand{\ie}{\textit{i.e}\xperiod}

\usepackage{bm}

\definecolor{cvprblue}{rgb}{0.21,0.49,0.74}
\usepackage[pagebackref,breaklinks,colorlinks,citecolor=cvprblue]{hyperref}

\def\paperID{*****} 
\def\confName{CVPR}
\def\confYear{2024}

\usepackage[capitalize]{cleveref}
\crefname{section}{Sec.}{Secs.}
\Crefname{section}{Section}{Sections}
\Crefname{table}{Table}{Tables}
\crefname{table}{Tab.}{Tabs.}

\begin{document}

\title{Domain-Adaptive Full-Face Gaze Estimation via \\Novel-View-Synthesis and Feature Disentanglement}

\author{Jiawei Qin$^{1}$, Takuru Shimoyama$^{1}$, Xucong Zhang$^{2}$, Yusuke Sugano$^{1}$ \\
\normalsize $^1$ Institute of Industrial Science, The University of Tokyo, Komaba 4-6-1, Tokyo, Japan \\
\normalsize $^2$ Computer Vision Lab, Delft University of Technology, Mekelweg 5, Delft, Netherlands \\
{
\tt\small \{jqin, tshimo, sugano\}@iis.u-tokyo.ac.jp} \\
\tt\small xucong.zhang@tudelft.nl
}
\maketitle

\begin{abstract}
Along with the recent development of deep neural networks, appearance-based gaze estimation has succeeded considerably when training and testing within the same domain.
Compared to the within-domain task, the variance of different domains makes the cross-domain performance drop severely, preventing gaze estimation deployment in real-world applications.
Among all the factors, ranges of head pose and gaze are believed to play significant roles in the final performance of gaze estimation, while collecting large ranges of data is expensive. 
This work proposes an effective model training pipeline consisting of a training data synthesis and a gaze estimation model for unsupervised domain adaptation.
The proposed data synthesis leverages the single-image 3D reconstruction to expand the range of the head poses from the source domain without requiring a 3D facial shape dataset.
To bridge the inevitable gap between synthetic and real images, we further propose an unsupervised domain adaptation method suitable for synthetic full-face data.
We propose a disentangling autoencoder network to separate gaze-related features and introduce background augmentation consistency loss to utilize the characteristics of the synthetic source domain.
Through comprehensive experiments, it shows that the model using only our synthetic training data can perform comparably to real data extended with a large label range.
Our proposed domain adaptation approach further improves the performance on multiple target domains.
The code and data will be available at \url{https://github.com/ut-vision/AdaptiveGaze}.
\end{abstract}

\section{Introduction}

\begin{figure}[t]
\begin{center}
  \includegraphics[width=0.99\linewidth]{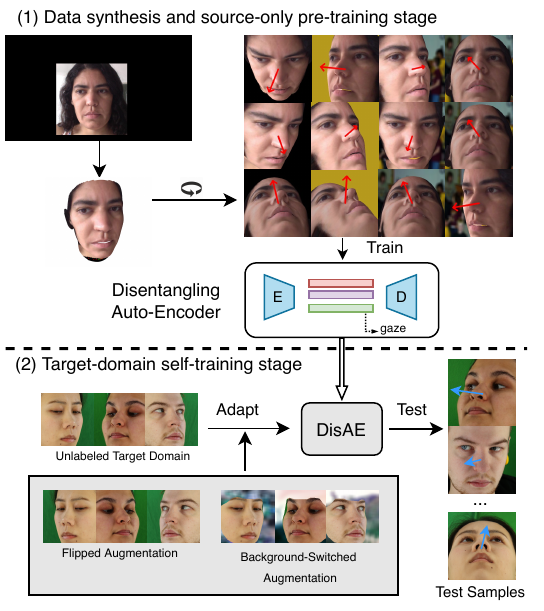}
\end{center}
  \caption{Overview of our approach with two stages.
  (1) With a monocular source image as input, we synthesize the data to be a large range of head poses stemming from the 3D face reconstruction. We propose a feature disentangling auto-encoder network pre-trained only on the synthetic data from the source images.
  (2) For the unlabeled target domain,  We leverage self-training to adapt the model to unlabeled target domains. 
  } 
\label{fig:teaser}
\end{figure}

Appearance-based gaze estimation is a promising solution to indicate human user attention in various settings with a single webcam as the input device, such as human-robot interaction~\cite{majaranta2014eye,mutlu2009footing}, social interaction~\cite{emery2000eyes,holzman1974eye}, and entertainment~\cite{corcoran2012real,scalera2021human}. 
Machine learning-based methods have been evolving from convolutional neural networks~\cite{Park2018ECCV,swcnn_zhang2017s,Zhang2020ETHXGaze} to vision transformer (ViT)~\cite{ViT_dosovitskiy2021an,cheng2022gazetr} for more robust performance under the in-the-wild usage setting. 
There is also a trend to include the face image instead of just the eye region \cite{swcnn_zhang2017s,gazecapture}.
Despite their effectiveness, it is well known that data-driven methods are prone to be overfitted to data bias to lose their generalization ability across environments, thus, are limited in real-world applications.

In the research community, this issue has often been analyzed quantitatively in cross-domain evaluations with training and testing on different datasets~\cite{mpii_zhang19_pami}.
There is a significant performance drop if the trained model is tested on different domains regarding personal appearances, head poses, gaze directions, and lighting conditions \cite{gaze360_2019,Zhang2020ETHXGaze,wang2019generalizing}.
To improve the robustness of the deep estimation model, many large-scale datasets have been collected and contributed to the community~\cite{Zhang2020ETHXGaze,mpii_zhang19_pami,eyediap_Mora_ETRA_2014,gazecapture,gaze360_2019}. 
In-the-wild setting datasets achieved diverse lighting~\cite{mpii_zhang19_pami,gaze360_2019} and large subject scale~\cite{gazecapture}, while controlled settings can capture images with extreme head pose and gaze direction~\cite{Zhang2020ETHXGaze}.
However, acquiring gaze labels is costly, and constructing a comprehensive training dataset covering all conditions is not trivial.

Another line of work created synthetic data as additional training data~\cite{yu2019_syn_user_specific,sted_Zheng2020NeurIPS,learn_by_syn_Sugano_2014_CVPR,gazeonce_ZhangCVPR22}.
Although synthetic images from gaze redirection can be used to augment existing training data~\cite{sted_Zheng2020NeurIPS}, the label accuracy from the learning-based generative model is not good enough as training data alone, especially for large angles.
Learning-by-synthesis using 3D graphics has been proven effective in eye-only gaze estimation~\cite{yu2019_syn_user_specific,learn_by_syn_Sugano_2014_CVPR,wood2016_etra}. 
However, the domain gap between real and synthetic images is difficult to fill~\cite{simgan17_CVPR}, and no effective pipeline has been proposed to synthesize and adapt full-face images as training data.

In this work, we propose tackling the goal of generalizable appearance-based gaze estimation by leveraging data synthesis and the domain adaptation approach. 
As shown in the top of \Fref{fig:teaser}, we perform single-image 3D face reconstruction to synthesize data for large head poses and extend the gaze direction ranges.
With the synthetic data, we propose a novel unsupervised domain adaptation framework combining disentangled representation learning and a self-training strategy. 
Our proposed disentangling auto-encoder (\methodname) structure is first trained on the synthetic source domain for learning gaze representation expected to better generalize to unseen domains.
The model is then trained on unlabeled target domains in a self-training manner~\cite{amini2022self, xie2020unsupervised, ohkawa_access21_style, ohkawa_eccv22_DAhand, liu2021PnP_GA}.
Based on the characteristics of our synthetic data, we propose to use background-switching data augmentation consistency loss for the synthetic-real domain adaptation.
Experiments with multiple target datasets show that the proposed pipeline significantly improves performance from the source dataset before reconstruction.
Our single-image face reconstruction approach applies to most real-world settings, yet the multi-view face reconstruction could pose an upper bound in performance.
We also analyze in detail how far single-view reconstruction can approach the training accuracy of multi-view reconstruction.

This manuscript is based on our previous publication~\cite{qin2022learning}, and parts of the text and figures are reused from the previous version. 
The major changes are as follows.
First, we fully update the model training pipeline by introducing \methodname and a self-training strategy for unsupervised domain adaptation. 
Second, we added an experiment using multi-view reconstruction data from the ETH-XGaze datasets~\cite{Zhang2020ETHXGaze} to analyze the upper-bound performance of synthetic training.
This results in re-calibrated camera parameters for the ETH-XGaze dataset, promoting future multi-view gaze estimation tasks.

In summary, the contributions of this work are threefold.
\begin{enumerate} [label=(\roman*)]
\item We are the first to propose a novel approach using single-view 3D face reconstruction to create synthetic training data for appearance-based gaze estimation. 
We utilize the property of the synthetic data to perform the background switching for image appearance argumentation.
\item We propose a novel unsupervised domain adaptation approach combining feature disentanglement and self-training strategy.
Experiments show that the proposed method is particularly effective in addressing synthetic-real domain gaps.
\item We provide a detailed comparison with multi-view face reconstruction to analyze the single-view performance. 
We release the re-calibrated camera extrinsic parameters for the ETH-XGaze dataset to facilitate further research.
\end{enumerate}


\section{Related works}

\subsection{Appearance-Based Gaze Estimation}
While traditional model-based gaze estimation relies on 3D eyeball models and geometric features~\cite{model_based1,model_2_hansen2009eye}, appearance-based methods~\cite{Tan2002AppearancebasedEG} use a direct mapping from the image to the gaze direction, enabling their use in a wider range of settings and with less hardware dependency.
Previous work on appearance-based gaze estimation has mostly used single eye~\cite{eye_2007,Tan2002AppearancebasedEG,learn_by_syn_Sugano_2014_CVPR, zhang15_cvpr, xiong2019mixed, yu2019_syn_user_specific, yu2020unsupervised} or two-eye images as the input~\cite{eye_asym_2020, Cheng_2018_ECCV, faze_Park2019ICCV, dagen_guo2020}. 
Recent works using full-face input~\cite{gazecapture, swcnn_zhang2017s, region_selection_Zhang2020, cheng2020coarse} have shown higher robustness and accuracy of gaze estimation than the eye-only methods.

To alleviate the data hunger of deep learning methods, multiple datasets have been proposed.
Most of the gaze datasets usually collected the data in indoor environments that lack variant lighting conditions \cite{colombia_CAVE_0324,FunesMora_ETRA_2014,rtgene_Fischer_2018_ECCV,huang2017tabletgaze}.
Later works switched to \textit{in-the-wild} data collection to cover variant lighting conditions \cite{mpii_zhang19_pami, swcnn_zhang2017s}.
However, these datasets have limited ranges of head pose and gaze directions due to the data collection devices such as the laptop~\cite{swcnn_zhang2017s, mpii_zhang19_pami}, cellphone \cite{gazecapture}, and tablet~\cite{gazecapture,huang2017tabletgaze}.
Recent datasets have further extended diversity in head pose and environment conditions~\cite{Zhang2020ETHXGaze,gaze360_2019}.
However, acquiring training datasets that meet the requirements for head pose and appearance variations in the deployment environment still requires significant effort.
In this work, we extend the head pose ranges of source datasets with full-face synthetic data for the gaze estimation task.

\subsection{Learning-by-Synthesis for Gaze Estimation}
Previous studies have created synthetic training data for the gaze estimation task to bypass the burden of real-world data collection.
One direction is to use multi-view stereo reconstruction~\cite{learn_by_syn_Sugano_2014_CVPR}. 
However, the multi-view setup has the drawback that the environment is limited to the laboratory conditions. 
Another group of methods used hand-crafted computer graphics models to generate the samples with arbitrary head poses, gaze directions, and lighting conditions~\cite{egp.20161054, wood2016_etra}. 
Unfortunately, these generated samples from the graphics models have a non-negligible domain gap between the synthesis and realisim. 
Gaze redirection has been proposed to generate synthetic data for the personal gaze estimator training~\cite{sted_Zheng2020NeurIPS,he2019GAN_eye,yu2019_syn_user_specific}.
However, these approaches cannot guarantee that the generated samples have exactly the target gaze label.
Alternatively, this work uses a single-image 3D face reconstruction approach for accurate data synthesis, enabling us to generate synthetic training data with higher realism and precision than previous methods.

\subsection{Domain Gap in Gaze Estimation}

The cross-domain gap is a significant challenge in appearance-based gaze estimation, and it becomes more critical with synthetic data.
To tackle this, previous works either improved the generalizability by devising better gaze representation learning from the source domain~\cite{faze_Park2019ICCV,cheng2022puregaze, yu2020unsupervised} or directly used target domain data with unsupervised domain adaptation~\cite{liu2021PnP_GA, dagen_guo2020, lee2022latentgaze}.
For instance, disentangling transforming encoder-decoder~\cite{faze_Park2019ICCV} separates the features to get more domain-variant gaze features.
PureGaze~\cite{cheng2022puregaze} extracts the purified gaze features out of the image feature using a self-adversarial learning strategy.
However, domain generalization in gaze estimation remains challenging due to numerous influencing factors.
This study is intended to synthesize data tailored to the head pose distribution of the target domain and primarily consider adaptation rather than generalization.

Unsupervised domain adaptation has succeeded in tasks like classification and segmentation~\cite{gao2022cross, huang2022category, xiao20233d}, but only limited work has focused on regression tasks~\cite{chen2021representation, nejjar2023dare, ohkawa_eccv22_DAhand, ohkawa_access21_style}, of which gaze estimation is particularly challenging.
SimGAN~\cite{simgan17_CVPR} adapts the synthetic training data to be similar to real target images before training, while recent methods are focusing more on directly adapting the model by target domain using self-supervised learning~\cite{jaiswal2020survey, khan2022contrastive, he2020momentum, lee2022latentgaze}.
Liu~\etal~\cite{liu2021PnP_GA} proposed a framework using collaborative learning that adapts to the target domain with only very few images guided by outlier samples.
Some methods pinpointed some specific issues such as jitter samples~\cite{liu2022jitter} and in-plane rotation inconsistency~\cite{Bao_2022_CVPR} and developed specific self-supervised learning strategies to address them.
GazeCLR~\cite{jindal2022gazeclr} leveraged multi-view consistency as the underlying principle of the proposed contrastive learning framework.
Wang~\etal~\cite{wang2022contrastive} proposed a contrastive learning framework based on an assumption of the similarity between gaze labels and features.
LatentGaze~\cite{lee2022latentgaze} leveraged generative adversarial networks to transform the target domain to the source domain for easier estimation.
In summary, most of the previous work only focused on adapting a wide range to a narrow range, and the effectiveness on synthetic source data has not been evaluated.
Thus, we propose a gaze estimation model that specifically learns better gaze representation from synthetic data and adapts to the real domain using unlabeled data.

\begin{figure*}[ht]
\begin{center}
  \includegraphics[width=0.88\linewidth]{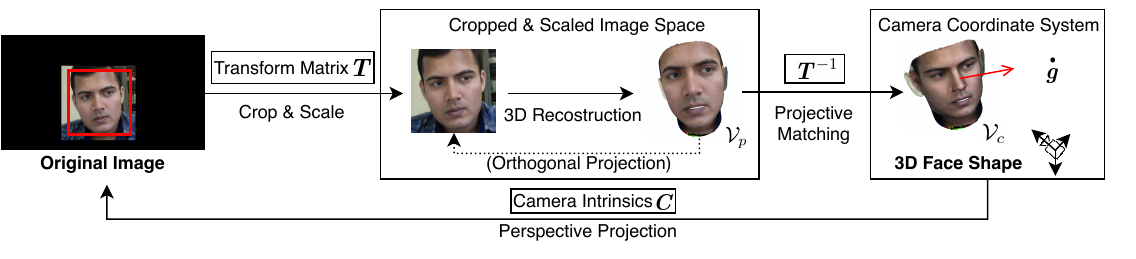}
\end{center}
  \caption{Overview of the data synthesis pipeline. 
  With monocular image as input, we first obtain the face patch with croping and scaling. 
  We then fit the 3D face model to the input face patch. 
  We assume that 3D face reconstruction methods generate facial meshes under an orthogonal projection model. 
  Through the proposed projective matching, we convert the mesh from the image-pixel system to the physical camera coordinate system. 
  After this process, the 3D face is aligned with the ground-truth gaze position (in the physical camera coordinate system), thus we can rotate the 3D face to simulate different head poses. 
  }

\label{fig:overview}
\end{figure*}

\subsection{3D Face Reconstruction}

There has been significant progress in the monocular 3D face reconstruction techniques in recent years~\cite{Zollhfer2018StateOT}. 
Despite the fact that reconstructed 3D faces have also been used to augment face recognition training data~\cite{rec_recog_2004,zhou2020rotate,7961797}, no prior work has explored its usage in full-face appearance-based gaze estimation yet.
Most of the methods based on 3D morphable models~\cite{deng2019accurate,Tran_2017_CVPR} approximate facial textures via the appearance basis~\cite{3dmm,bfm09,FLAME:SiggraphAsia2017} that the appearances of the eye region can be distorted. 
To preserve accurate gaze labels after reconstruction, the proposed data synthesis approach utilizes 3D face reconstruction methods that sample texture directly from the input image~\cite{6412675, Zhu_2016_CVPR, bulat2017far, guo2020towards, Yao2021DECA, 3ddfa_cleardusk, zhu2017face}. 
In addition, since many prior works rely on orthogonal or weak perspective projection models, we also investigate how to precisely align the reconstruction results with the source camera coordinate system.

In addition to the monocular-based methods, there are multi-view stereo methods that produce a better 3D geometry~\cite{kaya2022neural} and Neural Radiance Field (NeRF) that represents the scene implicitly as a radiance field to achieve fine-level details. 
However, these methods require a long processing time and a large number of views, and they are sensitive to low-quality images~\cite{rosu2022neuralmvs}.


\section{Novel-View Synthesis}\label{sec:data_synthesis} 

Our proposed method consists of two parts: data synthesis and synthetic-real domain adaptation.
In this section, we introduce the data synthesis process with 3D face reconstruction to generate samples with a large range of head poses.

\subsection{Overview}

\Fref{fig:overview} shows the overview of our data synthesis pipeline. Given an ordinary single-view gaze dataset, we apply 3D face reconstruction on each sample to synthesize face images with novel head poses while preserving accurate gaze direction annotations.
We adopt a simple 3D face reconstruction to create the 3D face mesh in the pixel coordinate system, and propose a transformation process, named \textit{projective matching}, to obtain the 3D face mesh in the camera coordinate system.
Finally, 2D face images can be rendered using camera perspective projection with the 3D face mesh.

\subsection{3D Face Reconstruction}
The synthesis image generation requires the source gaze dataset consists of 1) face images, 2) the projection matrix (intrinsic parameters) $\bm{C}$ of the camera, and 3) the 3D gaze target position $\bm{g} \in \mathbb{R}^3$ in the camera coordinate system.
Most of the existing gaze datasets satisfy our requirements~\cite{zhang15_cvpr, gazecapture, Zhang2020ETHXGaze}, and yaw-pitch annotations can also be converted assuming a distance to the dummy target \cite{gaze360_2019}.
State-of-the-art learning-based 3D face reconstruction methods usually take a cropped face patch as input and output a 3D facial mesh, which is associated with the input image in an orthographic projection way.
Without loss of generality, we assume that the face reconstruction method takes a face bounding box defined with center $(c_x, c_y)$, width $w_b$, and height $h_b$ in pixels and then resized to a fixed input size by factor $(s_x, s_y)$.
The reconstructed facial mesh is defined as a group of $N$ vertices $\mathcal{V}_{p}=\{\bm{v}_p^{(i)}\}_{i=0}^N$.
Each vertex is represented as $\bm{v}_p^{(i)} = [u^{(i)},v^{(i)},d^{(i)}]^\top$ in the right-handed coordinate system, where $u$ and $v$ directly correspond to the pixel locations in the input face patch and $d$ is the distance to the $u$-$v$ plane in the same pixel unit. 
This representation has been used by recent works~\cite{3ddfa_cleardusk, feng2018prn, Zhu_2016_CVPR, Jourabloo_2015_ICCV, bulat2017far}, and we can convert arbitrary 3D representation to it by projecting the reconstructed 3D face onto the input face patch.

Our goal is to convert the vertices of the reconstructed 3D face $\mathcal{V}_{p}$ to another 3D representation $\mathcal{V}_{c}=\{\bm{v}_c^{(i)}\}_{i=0}^N$ where each vertex $\bm{v}_c^{(i)} = [x^{(i)},y^{(i)},z^{(i)}]^\top$ is in the original camera coordinate system so that it can be associated with the gaze annotation $\bm{g}$.
In this way, the gaze target location can also be represented in the facial mesh coordinate system, and we can render the facial mesh under arbitrary head or camera poses together with the ground-truth gaze direction information.

\subsection{Projective Matching}\label{projective-matching}
Projective matching, in a nutshell, is to approximate parameters for transforming the $\mathcal{V}_{p}$ to $\mathcal{V}_{c}$ such that $\mathcal{V}_{c}$ matches the perspective projection.

In detail, since $u$ and $v$ of each reconstructed vertex $\bm{v}_{p}$ are assumed to be aligned with the face patch coordinate system, $\bm{v}_{c}$ must be on the back-projected ray as 
\begin{equation}\label{eq:invproj}
\bm{v}_{c} =
\lambda \frac{\bm{C}^{-1} \bm{p}_o}{||\bm{C}^{-1} \bm{p}_o||} =
\lambda \frac{\bm{C}^{-1} \bm{T}^{-1} \bm{p}}{||\bm{C}^{-1} \bm{T}^{-1} \bm{p}||},
\end{equation}
where $\bm{p}_o = [u_o, v_o, 1]^{\top}$ and $\bm{p} = [u, v, 1]^{\top}$ indicates the pixel locations in the original image and the face patch in the homogeneous coordinate system, respectively, and
\begin{equation}
\bm{T} = \begin{bmatrix} s_x & 0  & - s_x (c_x - \frac{w}{2}) \\0 & s_y & -s_y (c_y - \frac{h}{2}) \\0 & 0 &1  \end{bmatrix}
\end{equation}
represents the cropping and resizing operation to create the face patch, \ie, $\bm{p} = \bm{T}\bm{p}_o$.
The scalar $\lambda$ indicates scaling along the back-projection ray and physically means the distance between the camera origin and $\bm{v}_{c}$.

\begin{figure}[t]
\begin{center}
  \includegraphics[width=0.7\linewidth]{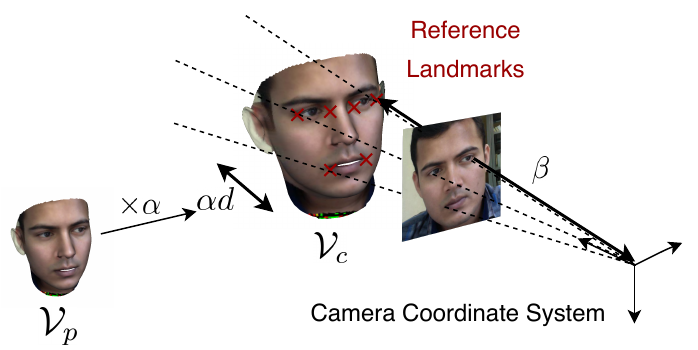}
\end{center}
  \caption{Determining the location of $\mathcal{V}_{c}$ via parameters $\alpha$ and $\beta$. $\alpha$ indicates a scaling factor from the pixel to physical (\eg, millimeter) unit, and $\beta$ is the bias term to align $\alpha d$ to the camera coordinate system.}
\label{fig:alpha-beta}
\end{figure}

Since Eq.~(\ref{eq:invproj}) does not explain anything about $d$, our task can be understood as finding $\lambda$, which also maintains the relationship between $u$, $v$, and $d$.
Therefore, as illustrated in \Fref{fig:alpha-beta}, we propose to define $\lambda$ as a function of $d$ as $\lambda = \alpha d + \beta$.
$\alpha$ indicates a scaling factor from the pixel to physical (\eg, millimeter) unit, and $\beta$ is the bias term to align $\alpha d$ with the camera coordinate system.
Please note that $\alpha$ and $\beta$ are constant parameters determined for each input image and applied to all vertices from the same image.

We first fix $\alpha$ based on the distance between two eye centers (midpoints of two eye corner landmarks) compared to a physical reference 3D face model.
3D face reconstruction methods usually require facial landmark detection as a pre-processing step. 
Thus we can naturally assume that the corresponding vertices in $\mathcal{V}_{p}$ to the eye corner landmarks are known.
We use a 3D face model with 68 landmarks (taken from the OpenFace library~\cite{openface}) as our reference.
We set $\alpha = l_{r}/l_{p}$, where $l_{p}$ and $l_{r}$ are the eye-center distances in $\mathcal{V}_{p}$ and in the reference model, respectively.

We then determine $\beta$ by aligning the reference landmark depth in the camera coordinate system. 
In this work, we use the face center as a reference, which is defined as the centroid of the eyes and the mouth corner landmarks, following previous works on full-face gaze estimation~\cite{swcnn_zhang2017s, zhang18_etra}. 
We use the same face center as the origin of the gaze vector through the data normalization and the gaze estimation task.

We approximate $\beta$ as the distance between the ground-truth 3D reference location and the scaled/reconstructed location as $\beta = ||\bar{\bm{v}}|| - \alpha \bar{d}$.
$\bar{d}$ is the reconstructed depth values computed as the mean of six landmark vertices corresponding to the eye and mouth corner obtained in a similar way as when computing $\alpha$.
$\bar{\bm{v}}$ is the centroid of the 3D locations of the same six landmarks in the camera coordinate system, which are obtained by minimizing the projection error of the reference 3D model to the 2D landmark locations using the Perspective-n-Point (PnP) algorithm~\cite{pnp_proj}.

\subsection{Training Data Synthesis}\label{rotate-and-render}

\begin{figure*}[t]
\begin{center}
  \includegraphics[width=0.98\linewidth]{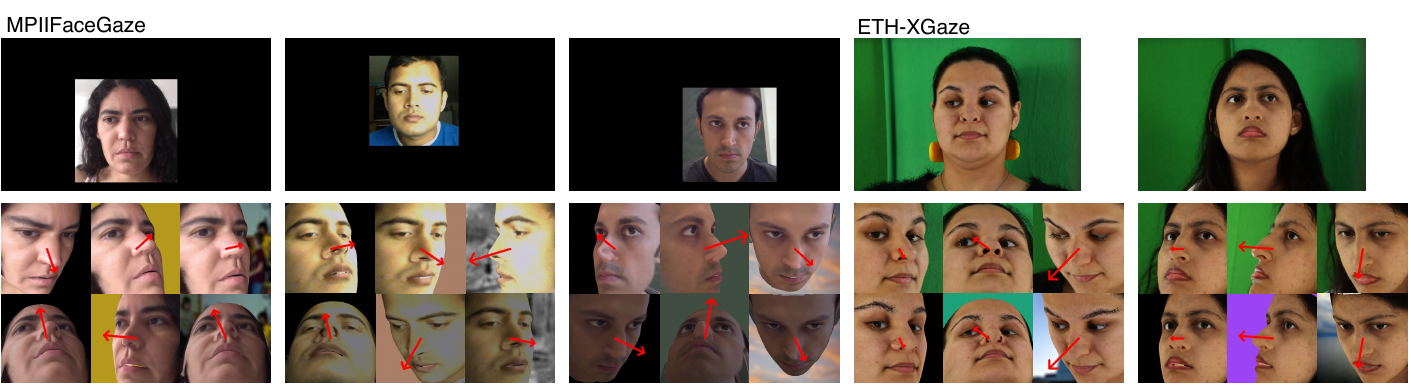}
\end{center}
  \caption{Examples of the synthesized images. The first row shows the source images from MPIIFaceGaze~\cite{swcnn_zhang2017s} and ETH-XGaze~\cite{Zhang2020ETHXGaze} datasets. 
  For MPIIFaceGaze, the second and third rows show synthesized images in full and weak lighting. 
  For ETH-XGaze, the second row shows the real images from the dataset, and the third row shows our synthetic images with the same head poses as the real samples. 
  For each synthetic example, the three columns show the black, color, and scene background in turn. 
  The red arrows indicate gaze direction vectors.} 
\label{fig:render-example}
\end{figure*}

With the 3D face mesh $\mathcal{V}_{c}$ in the original camera coordinate system, we can render 2D face images under arbitrary head poses with the ground-truth gaze vector. To render a face image in a new camera coordinate system defined by the extrinsic parameters $\bm{R}_e, \bm{t}_e$, we project the vertex $\bm{v}_c$ and gaze target position $\bm{g}$ onto the new system by applying the transformation $\bm{R}_e \bm{v}_{c} + \bm{t}_e$ and $\bm{R}_e \bm{g} + \bm{t}_e$, respectively. 
To render a face image from a source head pose $\bm{R}_s, \bm{t}_s$ to a target head pose $\bm{R}_t, \bm{t}_t$, we transform the vertices and gaze position by applying the transformation $ \bm{R}_t (\bm{R}_s)^{-1} (\bm{v}_c -\bm{t}_s) + \bm{t}_t$, similar for $\bm{g}$.

Except for the geometric augmentation, we further augment the data with image appearances in terms of lighting conditions and background appearances by virtue of the flexible synthetic rendering.
Although most 3D face reconstruction methods do not reconstruct lighting and albedo, we maximize the diversity of rendered images by controlling the global illumination.
In the PyTorch3D renderer, the ambient color $[r,g,b]$ represents the ambient light intensity, ranging from $0$ to $1$, in which $1$ is the default value for full lighting.
For weak-light images, we set them to be a random value between $0.25$ and $0.75$.
We set the background to random colors or scenes by modifying the blending settings. 
Random scene images are taken from the Places365 dataset~\cite{zhou2017places}, and we apply blurring to them before rendering faces.
Overall, among all generated images, the ratio of black, random color, and random scene are set to 1:1:3, and half of them are weak lighting.
\Fref{fig:render-example} shows examples of the synthesized images using MPIIFaceGaze~\cite{mpii_zhang19_pami} and ETH-XGaze~\cite{Zhang2020ETHXGaze}.

In the experiments, we applied 3DDFA~\cite{3ddfa_cleardusk} to reconstruct 3D faces from the source dataset.
After projective matching, we rendered new images using the PyTorch3D library~\cite{ravi2020pytorch3d}.


\section{ Domain Adaptation with Feature Disentanglement}\label{sec:disae_da}

Our synthetic dataset generation pipeline can render realistic face regions with accurate gaze labels, and these generated samples could be directly used to train a model for the cross-domain task.
However, there is still an image appearance domain gap between the synthetic and real samples.
In particular, the influence of the background, hair, clothing, and other non-face areas of the synthesized images on the gaze estimation model cannot be ignored.
To fill the gap, we propose a gaze estimation framework that can adjust to the target domain by unsupervised domain adaptation.
Gaze-unrelated features, such as image appearance and head pose, make the adaptation unstable~\cite{qin2022learning,lee2022latentgaze}.
To avoid being disrupted by the unrelated features, we first devise the disentangling auto-encoder (\methodname) (Sec.~\ref{sec:disae}) to separate the gaze-related features during the supervising training with the synthetic data from the source domain.
Then, we further adapt the \methodname to the target domain in a self-training approach~\cite{ohkawa_access21_style, ohkawa_eccv22_DAhand}.
Since our synthetic source domain has random images as the background (Sec.~\ref{rotate-and-render}), we propose to use background-switching consistency loss on the target domain as one of the self-training objectives.

\begin{figure*}[t]
\centering
     \includegraphics[width=0.85\linewidth]{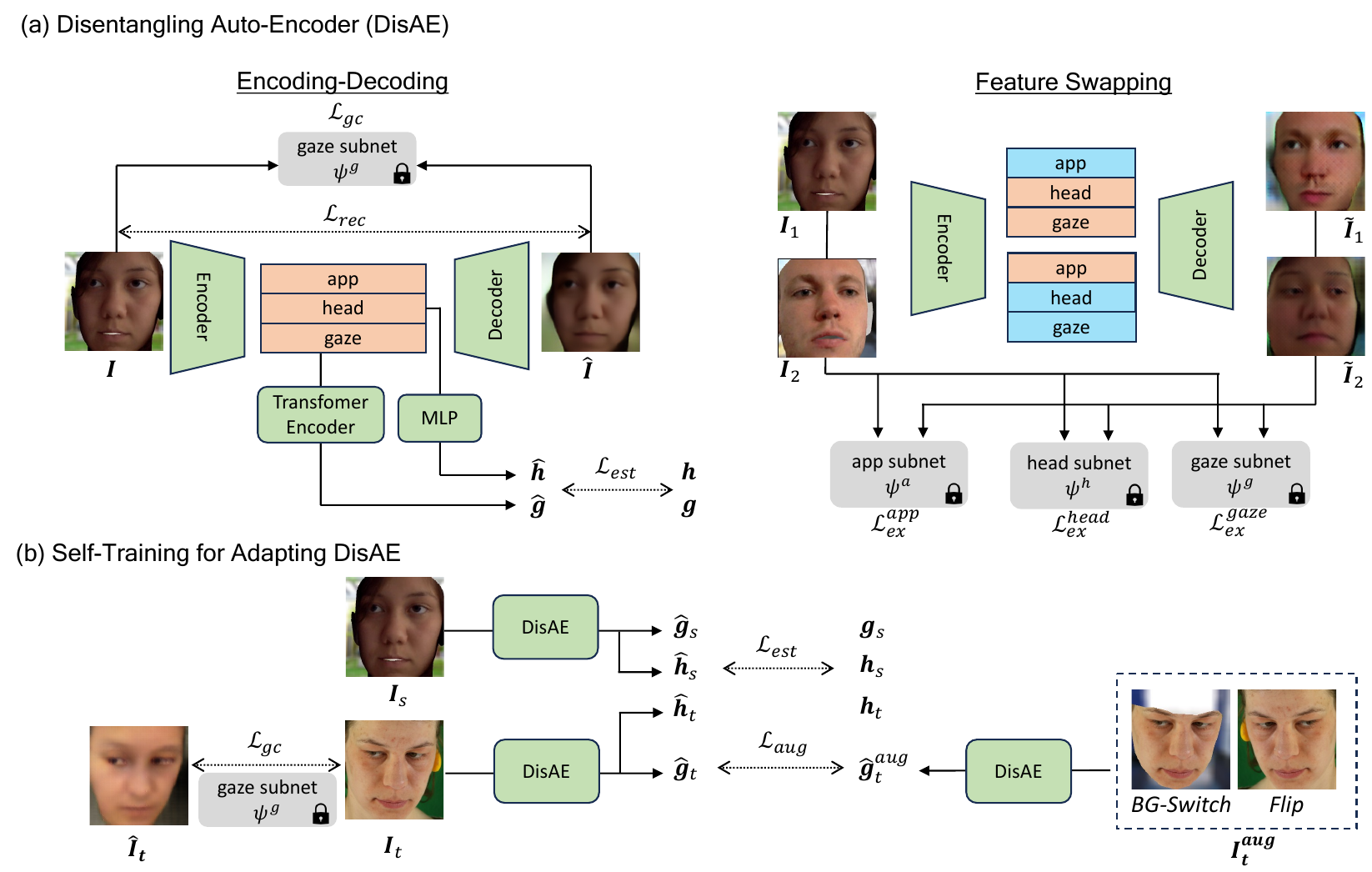}
     \label{fig:net1}
  \caption{The overview of our synthetic-real domain adaptation approach.
  Top: An encoder-decoder structure for feature disentanglement (\methodname).
  We prepare three subnets $\psi^{\textrm{a}}$, $\psi^{\textrm{h}}$, and $\psi^{\textrm{g}}$ to disentangle appearance, head and gaze features, respectively.
  The gaze features are fed into a vision transformer to get the predicted gaze direction $\hat{\bm{g}}$ and the head features are fed into an MLP to get the predicted head pose direction $\hat{\bm{h}}$.
  Bottom: augmentation consistency is proposed during the unsupervised domain adaptation of \methodname towards the target domain. 
  } 
\label{fig:network}
\end{figure*}

\subsection{Disentangling Auto-Encoder}\label{sec:disae}

As the base architecture for adaptation, we propose a disentangling auto-encoder to separate the gaze-unrelated features to reduce their influences on gaze estimation.
Specifically, as shown in the top of \Fref{fig:network}, we propose to use an encoder-decoder architecture to disentangle appearance, head, and gaze embeddings. For prediction, we use an MLP to predict head pose $\hat{\bm{h}}$, and a vision transformer to predict gaze direction $\hat{\bm{g}}$.
Finally, all features are concatenated and fed into a decoder to reconstruct the image. 
After this feature disentanglement, the extracted features are strongly correlated to gaze and ease the influence from other gaze-unrelated features, and the pre-trained model is expected to be more suited to domain adaptation.

To ensure the features are disentangled, we prepare three additional subnets denoted as face recognition network $\psi^{\textrm{a}}$ that predicts appearance embeddings, $\psi^{\textrm{h}}$ for predicting head pose, and $\psi^{\textrm{g}}$ for predicting gaze, all having a ResNet-18 structure.
The three subnets are trained on the source domain, and then we train the~\methodname~using the losses conducted by these three subnets.
Note that we only use the source domain synthetic data to pre-train the \methodname.
As shown in the top of \Fref{fig:network}, the pre-training loss mainly consists of two components, \textit{encoding-decoding} and \textit{feature swapping}.

\subsubsection*{Encoding-Decoding}

Encoding-decoding losses are defined between each input image and the output (decoded image and estimation result) from the \methodname architecture.
\textbf{Reconstruction loss} is defined as $\mathcal{L}_{\textrm{rec}} =  | \bm{I} - \hat{\bm{I}} |_1$, where $\hat{\bm{I}}$ is the reconstructed image and $\bm{I}$ is the ground-truth image. 
\textbf{Estimator loss} is the commonly used $\ell1$ loss on the gaze labels $\bm{g}$ defined as the pitch and yaw dimensions.
In addition to gaze loss, we will calculate the head pose loss with the dataset's head pose label $\bm{h}$. 
Taken together, the estimator loss is defined as
$\mathcal{L}_{\textrm{est}} =  | \bm{g} - \hat{\bm{g}} |_1 + \lambda_{head} | \bm{h} - \hat{\bm{h}} |_1 $,
where $\hat{\bm{g}}$ and $\hat{\bm{h}}$ are the predicted gaze and head direction.

\textbf{Gaze consistency loss} is aimed to make the gaze features not sensitive to different appearance features. 
We add an $\mathcal{N}(\mathbf{0}, \mathbf{0.1})$ random noise to the appearance features before feeding them into the decoder.
The reconstructed image $\hat{\bm{I}}$ with noise is expected to have the same gaze direction as the original image $\bm{I}$.
Therefore, we use the pre-trained gaze subnet $\psi^{g}$ to compute a gaze consistency loss between the two images as 
$\mathcal{L}_{\textrm{gc}} = | \psi^{g}(\bm{I}) - \psi^{g}(\hat{\bm{I}}) |_1 $.

\subsubsection*{Feature Swapping}

Feature swapping losses are defined between image pairs with disentangled features swapped after passing through the encoder.
\textbf{Feature exchange consistency loss} is introduced to enable the disentangling of the face appearances from the gaze and head pose features.
Specificially, we swap the appearance embeddings between two samples $\bm{I}_1$ and $\bm{I}_2$ and decode them to $\tilde{\bm{I}_1}$ and $\tilde{\bm{I}_2}$.
We ensure that $\tilde{\bm{I}_1}$ retains the head pose and gaze features of $\bm{I}_1$ while having the appearance features of $\bm{I}_2$.
Similarly, $\tilde{\bm{I}_2}$ retains the head pose and gaze features of $\bm{I}_2$ while having the appearance features of $\bm{I}_1$.
This process is illustrated in the top of \Fref{fig:network}, and the loss is formulated as 
\begin{equation}
    \mathcal{L}_{\textrm{ex}}(\bm{I}_1, \bm{I}_2) = \sum_{\psi}|\psi(\bm{I}_1) - \psi(\tilde{\bm{I}}_1) |_1 + |\psi(\bm{I}_2) - \psi(\tilde{\bm{I}}_2) |_1.
\end{equation}

The total loss for pre-training \methodname is the sum of all the above losses
\begin{equation}
    \mathcal{L}_{\textrm{source}} =  \mathcal{L}_{\textrm{est}} 
                        + \lambda_{\textrm{rec}} \mathcal{L}_{\textrm{rec}} 
                        + \lambda_{\textrm{gc}} \mathcal{L}_{\textrm{gc}} 
                        + \lambda_{\textrm{ex}} \mathcal{L}_{\textrm{ex}}.
\end{equation}

\subsection{ Self-Training on Target Domains }\label{sec:augloss}

After the source-domain-only supervised training of \methodname, we leverage the unlabeled target-domain data using an augmentation consistency loss, inspired by the wide use of data augmentation in self-training~\cite{ohkawa_access21_style, ohkawa_eccv22_DAhand, liu2021PnP_GA}.
Briefly, we apply data augmentations on unlabeled target domain images and enforce the model to output the same gaze direction for the original and augmented samples.
These data augmentations tune the gaze-related features in \methodname on the target domain without the gaze label.

The augmentation consistency loss is defined as the $\ell1$ loss between the gaze prediction of the original target image and that of the augmented target images as 
$\mathcal{L}_{\textrm{aug}}= \lambda_{\textrm{bg}} \mathcal{L}_{\textrm{bg}} + \lambda_{\textrm{flip}} \mathcal{L}_{\textrm{flip}}$, where 
$\mathcal{L}_{\textrm{bg}}=| \hat{\bm{g}}_{t} - \hat{\bm{g}}^{\textrm{bg}}_{t} |_1 $
and
$\mathcal{L}_{\textrm{flip}}=| \hat{\bm{g}}_{t} - \hat{\bm{g}}^{\textrm{flip}}_{t}  |_1$.   
As one of the data augmentation, we propose using a background-switching augmentation loss, particularly suitable for the synthetic data.
Specifically, we obtain the facial region mask from the detected landmarks and change the background to random images from Places365 dataset~\cite{zhou2017places}, which should not alter the gaze direction.
Since our synthetic source domain has random background regions, there is no gaze-relevant information in non-face regions and the pre-trained \methodname is expected to focus on face regions.
By training the model in the target domain so that gaze estimation remains consistent across images with swapped backgrounds, the model can be adapted to take advantage of this property.
Since most geometric augmentations may alter the results of gaze estimation, we only consider flipping as another data augmentation.
We flip the target image horizontally, negating the gaze label's yaw value.

In addition to the augmentation consistency loss, we use similar losses as the pre-training process.
We still compute the same gaze consistency loss $\mathcal{L}_{gc}$ for the target domain as 
$\mathcal{L}_{\textrm{gc}} = | \psi^{g}(\bm{I}_t) - \psi^{g}(\hat{\bm{I}_t}) |_1 $.
Since head pose labels can be obtained through the data normalization process even for the unlabeled target domain images, we define the estimator loss as
$\mathcal{L}_{\textrm{est}} =  | \bm{g}_s - \hat{\bm{g}_s} |_1 + \lambda_{head} ( | \bm{h}_s - \hat{\bm{h}_s} |_1 + | \bm{h}_t - \hat{\bm{h}_t} |_1 ) $, where $\hat{\bm{g}_s}$ and $\hat{\bm{h_s}}$ are the predicted gaze and head directions from source-domain input $\bm{I}_s$, respectively.
The $\hat{\bm{h}_t}$ is the predicted head direction of the target-domain input $\bm{I}_t$.

Consequently, the total loss for the adaptation stage is defined as
\begin{equation}\label{eq:loss_adapt}
    \mathcal{L}_{\textrm{adapt}} =  \mathcal{L}_{\textrm{est}}  
                        + \lambda_{\textrm{gc}} \mathcal{L}_{\textrm{gc}} 
                        + \mathcal{L}_{\textrm{aug}}. 
\end{equation}

\begin{figure*}[t]
\begin{center}
  \includegraphics[width=0.99\linewidth]{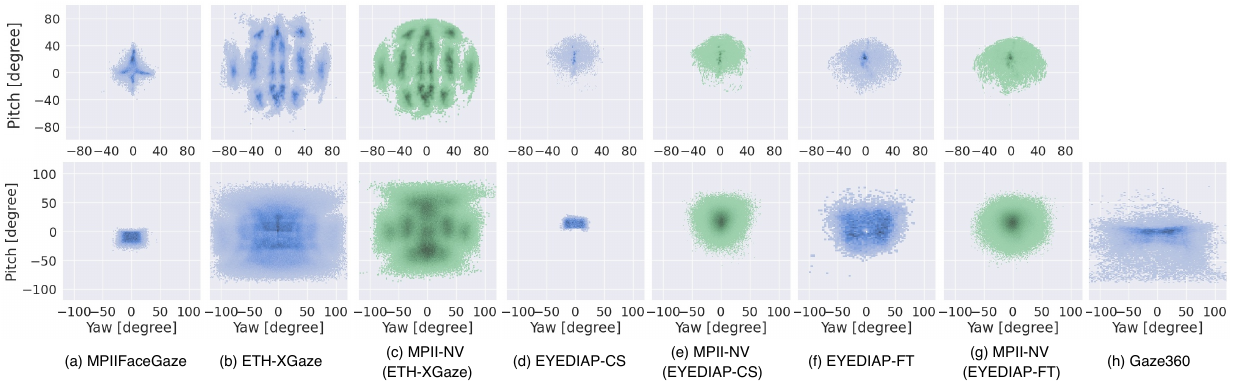}
\end{center}
  \caption{Head pose (top row) and gaze direction (bottom row) distributions of original datasets and our synthetic data. 
  We always take the MPIIFaceGaze (a) as the source dataset to extend for the distribution of another dataset. 
  By taking ETH-XGaze (b) as the target, we synthesize the data MPII-NV (ETH-XGaze) (c). 
  By taking EYEDIAP (CS) (d) as the target, we synthesize MPII-NV (EYEDIAP-CS) (e). 
  By taking EYEDIAP (FT) (f) as the target, we synthesize the MPII-NV (EYEDIAP-FT) (g). 
  The last column shows the gaze distribution of Gaze360 (h) as a reference, which does not provide the head pose distribution.
  }
\label{fig:distribution}
\end{figure*}

\subsection{Implementation Details}

The face recognition subnet $\psi^{a}$ is trained using a triplet loss~\cite{triplet1_wang2019multi} and the estimation subnets $\psi^{h}$ and $\psi^{g}$ are both trained using $\ell1$ loss.
For the source-only training stage for the \methodname, all subnets $\psi^{a}$, $\psi^{h}$, and $\psi^{g}$ are trained fully supervised with source-domain data.
For the target-domain adaptation stage, target-domain data is also used for fully supervised training of the face recognition subnet $\psi^{a}$ and head subnet $\psi^{h}$, while we use the same gaze subnet $\psi^{g}$ trained on the source domain.
We train the \methodname using the Adam optimizer~\cite{adam} for 12 epochs, setting the learning rate to 0.001, decaying to 0.1 every five epochs.
Apart from the target sample flipping in Sec~\ref{sec:augloss}, we also horizontally flip the images of the whole source training set to alleviate the inconsistent accuracy between horizontally symmetric images.

During adaptation, we randomly sample 2,000 samples from the target-domain datasets to adapt the model by ten epochs, and the final result is the average of five random-seed repetitions.
For the coefficients, we empirically set them to $\lambda_{\textrm{head}}=0.5$, $\lambda_{\textrm{rec}}=1.0$, $\lambda_{\textrm{ex}}=1.0$, $\lambda_{\textrm{gc}}=1.0$.
For the augmentation consistency losses, the \methodname in the bottom of \Fref{fig:network} share the weights.
Each type of augmentation loss is computed separately and added together, and we set $\lambda_{\textrm{bg}}=0.5$ and $\lambda_{\textrm{flip}}=0.5$.


\section{Experiments}

We conduct data extrapolation experiments (Sec.~\ref{subsec:data_extrap}) to demonstrate the angle extension of our proposed data synthesis and the adaptation of our proposed \methodname with ablation studies.
Additionally, we also compare multiple 3D face reconstruction methods in terms of their effects on the final gaze estimation performance as training data (Sec.~\ref{subsec:data_quality}).

\subsection*{Experimental Settings}

\paragraph{\textbf{Datasets}}
\textbf{MPIIFaceGaze}~\cite{swcnn_zhang2017s} 
consists of over 38,000 images of 15 subjects with variant lighting conditions. 
Since we only use this dataset for data synthesis, we selected images with frontal head poses that both pitch and yaw angles of the head pose are smaller than 15\textdegree.
To ensure that the number of samples from each subject is balanced for training, we randomly down-sampled or up-sampled the number of images of each subject to be 1,500. 
\textbf{ETH-XGaze}~\cite{Zhang2020ETHXGaze} contains over one million images of 110 subjects under variant head poses. 
We follow the official evaluation protocol and used the public evaluation server to retrieve the test results. 
\textbf{EYEDIAP}~\cite{eyediap_Mora_ETRA_2014} consists of over four hours of video data, using continuous screen targets (CS) or 3D floating object targets (FT). 
We treated the screen target and floating target subsets separately and sampled one image every five frames from the VGA videos using the pre-processing provided by Park~\etal~\cite{faze_Park2019ICCV}. 
\textbf{Gaze360}~\cite{gaze360_2019} consists of indoor and outdoor images of 238 subjects with wide ranges of head poses and gaze directions. 
We followed the pre-processing of Cheng~\etal~\cite{survey4_Cheng2021} which excluded cases with invisible eyes, resulting in 84,902 images.

We apply the data normalization scheme commonly used in appearance-based gaze estimation~\cite{zhang18_etra,Zhang2020ETHXGaze} for all datasets.
We also directly render the 3D facial mesh in the normalized camera space.
Unless otherwise noted, we follow the ETH-XGaze dataset~\cite{Zhang2020ETHXGaze} and set the virtual camera focal length to $960$ mm, and the distance from the camera origin to the face center to $300$ mm.
Face images are rendered in $448\times448$ pixels and down-scaled to $224\times224$ pixels before being fed into CNNs.
3D head pose is obtained by fitting a $6$-landmark 3D face model to the 2D landmark locations provided by the datasets, using the PnP algorithm~\cite{pnp_proj}.

\begin{table*}[t]
\caption{
Comparison of gaze estimation errors in degree. 
From left to right columns list the training sets, model architectures, and test sets. 
All models trained on our synthesized dataset \mpiitarget achieve the best performances on both ETH-XGaze and EYEDIAP-CS compared to other training datasets.
}
\centering
\scalebox{0.99}{
\begin{tabular}{l|c|cccc}
\hline
\multirow{2}{8em}{Training Datasets} & \multirow{2}{3em}{Model}   & \multicolumn{2}{c}{ETH-XGaze}& \multicolumn{2}{c}{EYEDIAP~\cite{eyediap_Mora_ETRA_2014}} \\ 
& & Train & Test  &  CS & FT \\ 
\hline
MPIIFaceGaze~\cite{swcnn_zhang2017s}       & \multirow{4}{6em}{ResNet18~\cite{resnet_He}}      &   31.96   &   32.62   &   13.02   &   23.01   \\
ETH-XGaze Train~\cite{Zhang2020ETHXGaze}    &       &   -       &   -       &   9.81    &\textbf{13.81}   \\ 
Gaze360~\cite{gaze360_2019}                 &       &   17.17   &   17.55   &   9.67    &   16.04   \\
\mpiitarget                                 &       &\textbf{12.84}&\textbf{13.99}&\textbf{5.37}& 17.04   \\ 
\hline
MPIIFaceGaze~\cite{swcnn_zhang2017s}       & \multirow{4}{7em}{PureGaze18~\cite{cheng2022puregaze}}     &   32.30   &   32.82   &   12.09   &   22.65   \\
ETH-XGaze Train~\cite{Zhang2020ETHXGaze}    &     &   -       &   -       &   8.79    &\textbf{12.86}\\ 
Gaze360~\cite{gaze360_2019}                 &     &   16.72   &   17.06   &   7.61    &   13.59   \\
\mpiitarget                                 &     &\textbf{12.93}&\textbf{14.07}&\textbf{5.45}& 16.22   \\ 
\hline
MPIIFaceGaze~\cite{swcnn_zhang2017s}       & \multirow{4}{6.5em}{GazeTR18~\cite{cheng2022gazetr}}     &   29.62   &   30.16   &   14.21   &   24.41   \\
ETH-XGaze Train~\cite{Zhang2020ETHXGaze}    &      &   -       &   -       &   8.91    &\textbf{12.61}\\ 
Gaze360~\cite{gaze360_2019}                 &      &   16.41   &   16.91   &   8.23    &   13.08   \\
\mpiitarget                                 &      &\textbf{11.36}&\textbf{12.01}&\textbf{5.58}& 14.70   \\ 
\hline
\end{tabular}
}
\label{tab:result_extrapolation}
\end{table*}

\paragraph{\textbf{Baseline Methods}}
We compare our method with several state-of-the-art gaze estimation methods.
Besides the simple yet strong baseline \textbf{ResNet}~\cite{resnet_He}, the \textbf{Gaze-TR}~\cite{cheng2022gazetr} is one of the state-of-the-art backbone architectures for single-image gaze estimation. 
It first extracts the gaze features from ResNet and feeds the feature maps into a transformer encoder followed by an MLP to output the gaze directions. 
\textbf{PureGaze}~\cite{cheng2022puregaze} first extracts image features using a ResNet, followed by an MLP for gaze estimation and decoding blocks for image reconstruction. 
\textbf{PnP-GA}~\cite{liu2021PnP_GA} is a domain adaptation model using Mean-Teacher~\cite{mean_teacher_tarvainen2017} structure and can be applied on many existing structures.
We follow the original implementation which only uses 10 target samples for adaptation.
\textbf{DANN}~\cite{ganin2015unsupervised} includes a gradient reverse layer and a domain classifier on top of the backbone, forcing the model to learn invariant features from source and target domains.

\subsection{Data Extrapolation}\label{subsec:data_extrap}

We explore the most practical setting data extrapolation, \ie, an extension of head poses and gaze directions from small ranges with synthesis data samples.
We extend the source MPIIFaceGaze dataset to a similar head pose distribution as the target ETH-XGaze and EYEDIAP datasets, respectively.
Note that we use the training set of the ETH-XGaze as the target head poses distribution.
We use the head pose values obtained through the data normalization process, and each source image is reconstructed and rendered with 16 new head poses randomly chosen from the target dataset.
To avoid extreme profile faces with fully occluded eyes, we discarded the cases whose pitch-yaw vector's $\ell$2-norm is larger than 80\textdegree~during the data synthesis.
As a result, the MPIIFaceGaze is extended to three synthetic datasets ETH-XGaze, EYEDIAP CS, and EYEDIAP FT, with 360,000 images, respectively.
We refer to these datasets as \mpiitarget.

\begin{table}[t]
\caption{Comparision of our method with baseline methods. 
The top block are source-domain-only methods, and the bottom block are methods that utilize unlabeled target-domain data.
}
\centering
\scalebox{0.9}{
\begin{tabular}{c|cccc}
\hline
 Model & \multicolumn{2}{c}{ETH-XGaze}& \multicolumn{2}{c}{EYEDIAP} \\ 
 & Train & Test  &  CS & FT \\ 
\hline
 ResNet18~\cite{resnet_He}      &   12.84       &   13.99       &   5.37      &   17.04     \\ 
 PureGaze18~\cite{cheng2022puregaze}    &   12.93       &   14.07       &   5.45      &   16.22     \\ 
 Gaze-TR18~\cite{cheng2022gazetr}     &   11.36       &   12.01       &   5.58      &   14.70     \\ 
\methodname (ours) & \textbf{11.21} & \textbf{12.00} & \textbf{5.22} & \textbf{13.50}\\ 
\hline
Res18 + PnP-GA~\cite{liu2021PnP_GA}  &  12.43 & 15.33 & 4.78 & 17.00 \\
Res18 + DANN~\cite{ganin2015unsupervised}  & 15.32 & 15.51 & 6.93 & 15.00 \\
Res18 + aug & 15.13 & 16.82 &  5.45 & 16.29 \\
\methodname+ DANN~\cite{ganin2015unsupervised} & 11.13 & 11.92 & 5.20 & 13.50 \\
\methodname+ aug (ours) &  \textbf{10.99} &   \textbf{11.89} & \textbf{4.63}  & \textbf{12.69}  \\
\hline
\end{tabular}
}
\label{tab:result_extrapolation_2}
\end{table}

\subsubsection{Comparison of Datasets}
We first evaluate how our data synthesis approach improves performance compared to other baseline training datasets.
As a real-image baseline, we compare the Gaze360~\cite{gaze360_2019} dataset which covers a wide gaze range.
The head pose and gaze distributions of the source and target real datasets (blue) and the synthetic datasets (green) are shown in \Fref{fig:distribution}, together with the gaze distribution of Gaze360 (head pose is not provided).
Since we synthesize the data based on head pose distribution, it can be seen that the gaze distribution does not exactly match the target but only roughly overlaps.

The comparison of training data under several baseline models is presented in \Tref{tab:result_extrapolation} and numbers represent the angular error. 
We compare different SOTA models in terms of gaze estimation performances when training on different training sets. 
From the table, we can see all models trained on our synthesized dataset \mpiitarget, achieve the best performances on both ETH-XGaze and EYEDIAP-CS compared to other training datasets.
Note the \mpiitarget is purely an extension of the original MPIIFaceGaze with large head poses. 
The significant improvements from models trained on \mpiitarget over MPIIFaceGaze indicate that our synthetic data pipeline can produce useful data for cross-dataset training.
For the EYEDIAP-FT, better performance was obtained when using real data ETH-XGaze.
One hypothesis is that EYEDIAP FT has a larger offset between gaze and head pose due to the use of physical gaze targets, such that our data synthesized based on head pose cannot fully reproduce the target gaze distribution (\Fref{fig:distribution}).

\subsubsection{Comparison of Models}

With the synthetic data, we evaluate the proposed \methodname for both cross-dataset and unsupervised domain adaptation settings.
In \Tref{tab:result_extrapolation_2}, we fix the~\mpiitarget as a training dataset to compare \methodname with SOTA methods.
The top block of \Tref{tab:result_extrapolation_2} shows performances of cross-dataset evaluation without adaptation, \ie, all methods are only trained with source-domain samples. 
We can see that~\methodname outperforms the three baseline models across all test datasets, showing the advantage of the feature disentanglement even without domain adaptation. 
The bottom block of \Tref{tab:result_extrapolation_2} shows the domain adaptation with unlabeled samples from the target test sets.
We can observe that our proposed \methodname model successfully adapts to all target domains, showing superior estimation errors in the last row.

To evaluate the effectiveness of combining \methodname with our self-training strategy, we apply the same augmentation consistency adaptation to the Res-18 networks and refer to it as \textit{Res18 + aug} in \Tref{tab:result_extrapolation_2}. 
This baseline achieves worse results than the \methodname showing that the unsupervised domain adaptation is difficult to be handled with simple data augmentation due to the gaze-unrelated features that exist in face images. 
Conversely, our proposed \methodname effectively alleviates this issue by focusing on gaze-related features, enhancing the model's adaptability to the target domain.
Furthermore, we apply DANN on the~\methodname by feeding the disentangled gaze features into the gradient reverse layer and domain classifier for a domain classification loss.
As in \textit{\methodname+ DANN}, though the DANN does not show remarkable effects on most of the test sets, \methodname demonstrates more stable adapting performance compared to the basic ResNet.

\subsubsection{Ablation Studies}

We first conduct an ablation study on source domain data augmentation using image flipping.
As shown in the top block of \Tref{tab:abla2_model}, \methodname achieves lower error than the non-flipping setting on all test datasets, proofing that flipping (doubling) the training data is a valuable and simple approach to deal with the inconsistency in the symmetric images.

For the domain adaptation stage, we examine individual loss terms based on Eq.~\ref{eq:loss_adapt}.
Note that we separately explore the flipping augmentation ($\mathcal{L}_{\textrm{flip}}$) and background replacement augmentation ($\mathcal{L}_{\textrm{bg}}$).
From the table, we can see that all proposed losses can gradually improve the accuracy of most of the test datasets.
The $\mathcal{L}_{\textrm{gc}}$ causes a negative effect only on EYEDIAP FT, which might be caused by the low image resolution and the float point existence of the EYEDIAP FT dataset.

\begin{table}[t]
    \caption{Adaptation effect with respect to the individual loss. 
    }
    \label{tab:abla2_model}
    \centering
    \scalebox{0.9}{
    \begin{tabular}{c|cccc}
        \hline
        &  \multicolumn{2}{c}{ETH-XGaze}& \multicolumn{2}{c}{EYEDIAP}  \\
         & Train & Test  &  CS & FT \\ 
        \hline
        \methodname~w.o. flip   &   11.32 & 12.03 & 5.90  &  13.85    \\ 
        \methodname  &   11.21 & 12.00 & 5.22  &  13.50    \\
        \hline
        $\mathcal{L}_{\textrm{gc}}$  &   11.10 & 11.90 & 5.22  &  13.43    \\
        $\mathcal{L}_{\textrm{bg}}$ &   11.12 & 12.00 & 4.96  &  11.54    \\
        $\mathcal{L}_{\textrm{flip}}$ &   11.40 & 12.21 & 5.36  &  11.40    \\
        $\mathcal{L}_{\textrm{bg}}$+$\mathcal{L}_{\textrm{flip}}$ &   11.32 & 12.12 & 4.81  &  \textbf{11.34}    \\
        $\mathcal{L}_{\textrm{gc}}$+$\mathcal{L}_{\textrm{bg}}$ & 11.03 & 11.99 & 4.97 &  13.00  \\ 
        $\mathcal{L}_{\textrm{gc}}$+$\mathcal{L}_{\textrm{flip}}$ & 11.09 & 11.95 & 4.80 & 12.94    \\ 
        $\mathcal{L}_{\textrm{gc}}$+$\mathcal{L}_{\textrm{bg}}$+$\mathcal{L}_{\textrm{flip}}$ (ours) &   \textbf{10.99} & \textbf{11.89} & \textbf{4.63}  &  12.69    \\
        \hline
    \end{tabular}
    }
\end{table}

\subsection{Analysis of Data Quality}\label{subsec:data_quality}

In this section, we aim to compare the face reconstruction performance between our proposed single-view and a more complex multi-view method. 
The primary goal of this comparison is to establish an upper bound of performance when using synthetic data for training gaze estimation models, given that the multi-view synthetic data offers higher photorealism.
In addition, we compare multiple methods of single-view reconstructions to analyze the impact of reconstruction performance on model accuracy.
We reconstruct the frontal-camera images of ETH-XGaze~\cite{Zhang2020ETHXGaze} and rotate them exactly to the other cameras, resulting in the synthetic version of ETH-XGaze, denoted as \textit{XGazeF-NV}.

\subsubsection{Multi-view Reconstruction}\label{subsec:multiview_recon}

As ETH-XGaze~\cite{Zhang2020ETHXGaze} is a camera-synchronized dataset, we implement multi-view reconstruction using the Agisoft Metashape software~\cite{agisoft_metashape}.
Since reconstruction quality under dark environments drops extensively, we only reconstruct the full-light frames.

\begin{figure}[t]
\begin{center}
  \includegraphics[width=0.99\linewidth]{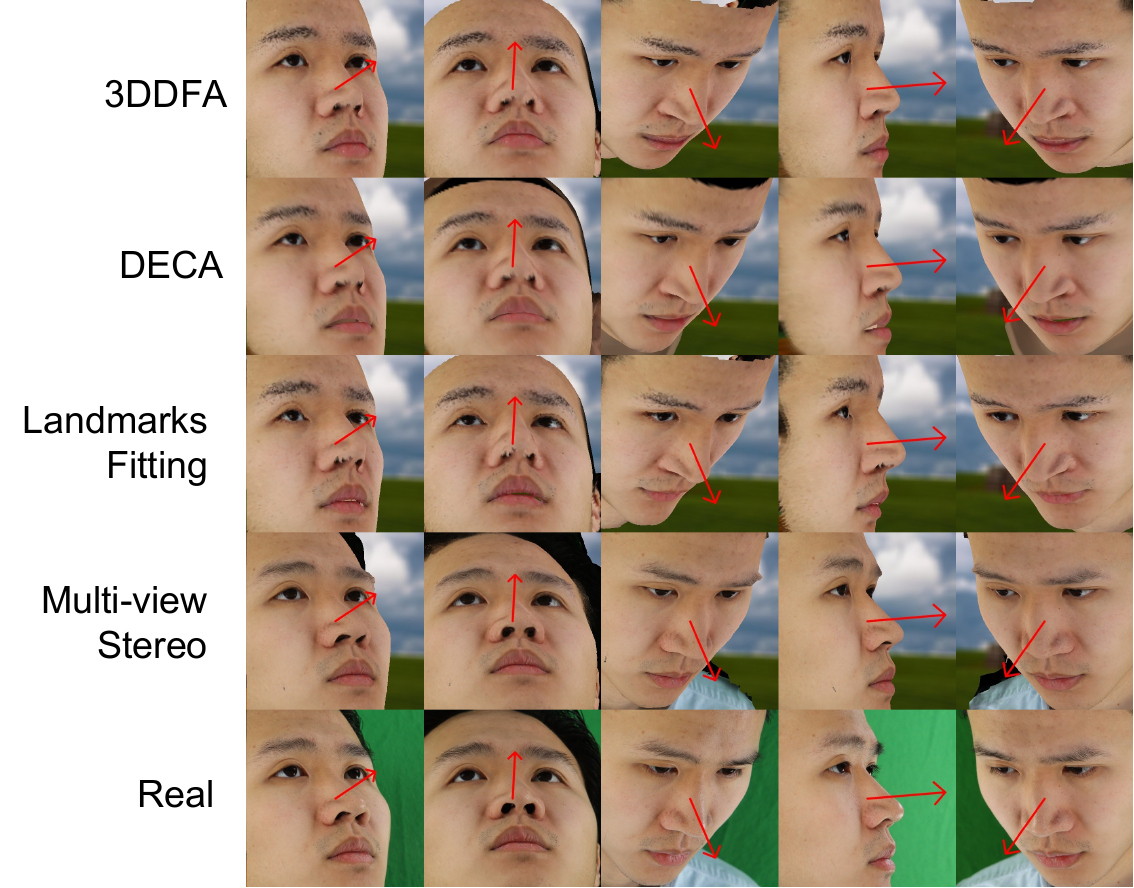}
\end{center}
   \caption{Examples of the synthesized XGaze-NV datasets using the different reconstruction methods.
   The last row shows the samples of the real dataset. 
    }
\label{fig:recon_compare}
\end{figure}

Through preliminary analysis, it is confirmed virtually that there is a discrepancy between the external parameters provided by the dataset and the actual camera images.
This is possibly due to the drift of the camera position after camera calibration, and the discrepancy varies for each subject.
Therefore, we first use Metashape to optimize the camera extrinsic parameters for each subject.
The Metashape optimization takes the 18 images of each frame of the target subject as input, as well as the fixed intrinsic parameters provided by the dataset.
This provides updated extrinsic parameters for each frame, and we discard frames whose camera position diverges by more than 10 mm from the raw average position.
We use the average value of the other remaining frames as the final extrinsic parameters for the subject.
We then use the recalibrated extrinsic parameters and the original intrinsic parameters to perform the multi-view face reconstruction.
Based on the reconstruction results, we render and denote the dataset as \mvxgaze.
Random factors for background augmentation are kept the same, and sample images from these reconstruction methods are shown in \Fref{fig:recon_compare}.

\subsubsection{Comparision of Reconstruction Methods}

We compare the multi-view face reconstruction with SOTA single-view methods 3DDFA~\cite{3ddfa_cleardusk} and DECA~\cite{Yao2021DECA}.
As another simplest baseline \textit{Landmarks Fitting}, we fit the BFM model~\cite{bfm09} to the detected 68 2D facial landmarks and get a facial shape, of which the texture is the RGB values obtained by projecting to the original image.

We separate the ETH-XGaze 80 training subjects into four folds to perform the leave-one-fold-out evaluation with a ResNet-18 model. 
We compare the performances of models trained on synthetic data generated by different face reconstruction methods.
The average errors are shown in \Tref{tab:recon_compare_bg_aug}.
From the top three rows block, we observe that the three single-view methods produce similar performance, and we attribute this to the similar appearance and texture (resolution) between all of these single-view methods.
On the other hand, as expected, the high-quality~\mvxgaze~shows the lowest error that is even close to the performance trained on real ETH-XGaze shown in the last two rows, which represents an upper bound of synthetic training.
Qualitatively, there are image quality differences between the single-view methods and \textit{Multi-view Stereo} in \Fref{fig:recon_compare} such as the artifacts in eyebrow, which may cause the above performance gap.

However, despite performing worse than the multi-view method, single-view methods are easier to deploy in real-world applications without complicated multi-view synchronization.
More importantly, there is a potential for reducing the gap between single-view and multi-view.
As shown in the fifth and the second row of \Tref{tab:recon_compare_bg_aug}, the proposed background augmentation reduces the error of \textit{XGazeF-NV-3DDFA}, which proves that appearance diversity is helpful for the generation of training data.
Besides the background augmentation, further refining the reconstruction quality, especially the texture, has the potential to further promote the single-view reconstruction methods closer to the performance of the upper-bound multi-view methods.

\begin{table}[t]
\caption{
Comparison of synthetic datasets using different reconstruction methods. 
The gaze estimation error is the average of the four-fold split.
The model used is ResNet18 for all rows.
$^\ast$ indicates the green background version without the background augmentation.
}
\label{tab:recon_compare_bg_aug}
\centering
\scalebox{0.9}{
\begin{tabular}{l|cc}\hline 
Datasets &  Within &  XGaze-Test  \\\hline
XGazeF-NV-3DDFA  &	10.03  & 11.32 \\
XGazeF-NV-DECA &   10.02 & 11.47\\
XGazeF-NV-fitting &	 10.28 &  11.67\\\hline
XGazeF-NV-3DDFA$^{\ast}$  &	20.33  & 22.85\\
\mvxgaze$^{\ast}$ &	 7.90 & 7.43\\
\mvxgaze &	 6.71 &  6.19\\
\hline
ETH-XGaze & 6.62 & 6.20 \\
\hline %
\end{tabular}
}
\end{table}

\section{Conclusion}
This work presents an effective data synthesis pipeline and an unsupervised domain adaptation approach for full-face appearance-based gaze estimation.
Our approach utilizes 3D face reconstruction to synthesize novel-head-poses training datasets while keeping accurate gaze labels via projective matching.
The proposed \methodname model can learn gaze-related features from the synthetic data and thus can effectively generalize to other domains. 
The \methodname can be further adapted to target domains through the self-training process using the proposed background-switching consistency loss.
Through extensive experiments, we show that the generated synthetic data can benefit the model training, and our approach achieves better performances than existing SOTA methods for cross-dataset and unsupervised domain adaptation evaluations.
Furthermore, experiments also verified that synthetic data can reach comparable performance as real data, pointing out the potential of synthetic training in future work.

\subsection*{Limitation}
Despite the effectiveness of our approach in extending the gaze range through novel view rendering using 3D reconstructed faces, our method has limitations. 
The proposed method does not fully explore image appearance augmentation, including variations in face size, textures, eye color, etc. 
This limitation may impact the generalization capabilities of baseline methods and also affect the disentangling capability of the proposed model. 
Therefore, future research should focus on synthesizing data that covers a broader spectrum of appearance, encompassing these variations. Addressing these factors will be crucial for improving the overall robustness and performance of gaze estimation models.

\subsection*{Acknowledgement}
This work was supported by JSPS KAKENHI Grant Number JP21K11932.

{\small
\bibliographystyle{ieeenat_fullname}
\bibliography{reference}

\begin{thebibliography}{93}
\providecommand{\natexlab}[1]{#1}
\providecommand{\url}[1]{\texttt{#1}}
\expandafter\ifx\csname urlstyle\endcsname\relax
  \providecommand{\doi}[1]{doi: #1}\else
  \providecommand{\doi}{doi: \begingroup \urlstyle{rm}\Url}\fi

\bibitem[Amini et~al.(2022)Amini, Feofanov, Pauletto, Devijver, and Maximov]{amini2022self}
Massih-Reza Amini, Vasilii Feofanov, Loic Pauletto, Emilie Devijver, and Yury Maximov.
\newblock Self-training: A survey.
\newblock \emph{arXiv preprint arXiv:2202.12040}, 2022.

\bibitem[Baltrusaitis et~al.(2018)Baltrusaitis, Zadeh, Lim, and Morency]{openface}
Tadas Baltrusaitis, Amir Zadeh, Yao~Chong Lim, and Louis-Philippe Morency.
\newblock Openface 2.0: Facial behavior analysis toolkit.
\newblock In \emph{Proc. FG}, 2018.

\bibitem[Bao et~al.(2022)Bao, Liu, Wang, and Lu]{Bao_2022_CVPR}
Yiwei Bao, Yunfei Liu, Haofei Wang, and Feng Lu.
\newblock Generalizing gaze estimation with rotation consistency.
\newblock In \emph{Proceedings of the IEEE/CVF Conference on Computer Vision and Pattern Recognition (CVPR)}, pages 4207--4216, 2022.

\bibitem[Belhumeur et~al.(2013)Belhumeur, Jacobs, Kriegman, and Kumar]{6412675}
Peter~N Belhumeur, David~W Jacobs, David~J Kriegman, and Neeraj Kumar.
\newblock Localizing parts of faces using a consensus of exemplars.
\newblock \emph{IEEE Trans. Pattern Anal. Mach. Intell.}, 35\penalty0 (12):\penalty0 2930--2940, 2013.

\bibitem[Blanz and Vetter(1999)]{3dmm}
Volker Blanz and Thomas Vetter.
\newblock A morphable model for the synthesis of 3d faces.
\newblock In \emph{Proc. SIGGRAPH}, 1999.

\bibitem[Bulat and Tzimiropoulos(2017)]{bulat2017far}
Adrian Bulat and Georgios Tzimiropoulos.
\newblock How far are we from solving the 2d \& 3d face alignment problem? (and a dataset of 230,000 3d facial landmarks).
\newblock In \emph{Proc. ICCV}, 2017.

\bibitem[Chen et~al.(2021)Chen, Wang, Wang, and Long]{chen2021representation}
Xinyang Chen, Sinan Wang, Jianmin Wang, and Mingsheng Long.
\newblock Representation subspace distance for domain adaptation regression.
\newblock In \emph{Proc. ICML}, pages 1749--1759, 2021.

\bibitem[Cheng and Lu(2022)]{cheng2022gazetr}
Yihua Cheng and Feng Lu.
\newblock Gaze estimation using transformer.
\newblock In \emph{Proc. ICPR}, 2022.

\bibitem[Cheng et~al.(2018)Cheng, Lu, and Zhang]{Cheng_2018_ECCV}
Yihua Cheng, Feng Lu, and Xucong Zhang.
\newblock Appearance-based gaze estimation via evaluation-guided asymmetric regression.
\newblock In \emph{Proc. ECCV}, 2018.

\bibitem[Cheng et~al.(2020{\natexlab{a}})Cheng, Huang, Wang, Qian, and Lu]{cheng2020coarse}
Yihua Cheng, Shiyao Huang, Fei Wang, Chen Qian, and Feng Lu.
\newblock A coarse-to-fine adaptive network for appearance-based gaze estimation.
\newblock In \emph{Proc. AAAI}, 2020{\natexlab{a}}.

\bibitem[Cheng et~al.(2020{\natexlab{b}})Cheng, Zhang, Lu, and Sato]{eye_asym_2020}
Yihua Cheng, Xucong Zhang, Feng Lu, and Yoichi Sato.
\newblock Gaze estimation by exploring two-eye asymmetry.
\newblock \emph{IEEE Trans. Image Process.}, 29:\penalty0 5259--5272, 2020{\natexlab{b}}.

\bibitem[Cheng et~al.(2021)Cheng, Wang, Bao, and Lu]{survey4_Cheng2021}
Yihua Cheng, Haofei Wang, Yiwei Bao, and Feng Lu.
\newblock Appearance-based gaze estimation with deep learning: A review and benchmark.
\newblock \emph{arXiv preprint arXiv:2104.12668}, 2021.

\bibitem[Cheng et~al.(2022)Cheng, Bao, and Lu]{cheng2022puregaze}
Yihua Cheng, Yiwei Bao, and Feng Lu.
\newblock Puregaze: Purifying gaze feature for generalizable gaze estimation.
\newblock \emph{Proc. AAAI}, 2022.

\bibitem[Corcoran et~al.(2012)Corcoran, Nanu, Petrescu, and Bigioi]{corcoran2012real}
Peter~M Corcoran, Florin Nanu, Stefan Petrescu, and Petronel Bigioi.
\newblock Real-time eye gaze tracking for gaming design and consumer electronics systems.
\newblock \emph{IEEE Transactions on Consumer Electronics}, 58\penalty0 (2):\penalty0 347--355, 2012.

\bibitem[Deng et~al.(2019)Deng, Yang, Xu, Chen, Jia, and Tong]{deng2019accurate}
Yu Deng, Jiaolong Yang, Sicheng Xu, Dong Chen, Yunde Jia, and Xin Tong.
\newblock Accurate 3d face reconstruction with weakly-supervised learning: From single image to image set.
\newblock In \emph{Proc. CVPRW}, 2019.

\bibitem[Dosovitskiy et~al.(2021)Dosovitskiy, Beyer, Kolesnikov, Weissenborn, Zhai, Unterthiner, Dehghani, Minderer, Heigold, Gelly, Uszkoreit, and Houlsby]{ViT_dosovitskiy2021an}
Alexey Dosovitskiy, Lucas Beyer, Alexander Kolesnikov, Dirk Weissenborn, Xiaohua Zhai, Thomas Unterthiner, Mostafa Dehghani, Matthias Minderer, Georg Heigold, Sylvain Gelly, Jakob Uszkoreit, and Neil Houlsby.
\newblock An image is worth 16x16 words: Transformers for image recognition at scale.
\newblock In \emph{Proc. ICLR}, 2021.

\bibitem[Emery(2000)]{emery2000eyes}
Nathan~J Emery.
\newblock The eyes have it: the neuroethology, function and evolution of social gaze.
\newblock \emph{Neuroscience \& biobehavioral reviews}, 24\penalty0 (6):\penalty0 581--604, 2000.

\bibitem[Feng et~al.(2018)Feng, Wu, Shao, Wang, and Zhou]{feng2018prn}
Yao Feng, Fan Wu, Xiaohu Shao, Yanfeng Wang, and Xi Zhou.
\newblock Joint 3d face reconstruction and dense alignment with position map regression network.
\newblock In \emph{Proc. ECCV}, 2018.

\bibitem[Feng et~al.(2021)Feng, Feng, Black, and Bolkart]{Yao2021DECA}
Yao Feng, Haiwen Feng, Michael~J. Black, and Timo Bolkart.
\newblock Learning an animatable detailed 3d face model from in-the-wild images.
\newblock \emph{TOG}, 40\penalty0 (4), 2021.

\bibitem[Fischer et~al.(2018)Fischer, Chang, and Demiris]{rtgene_Fischer_2018_ECCV}
Tobias Fischer, Hyung~Jin Chang, and Yiannis Demiris.
\newblock Rt-gene: Real-time eye gaze estimation in natural environments.
\newblock In \emph{Proc. ECCV}, 2018.

\bibitem[Fischler and Bolles(1981)]{pnp_proj}
Martin~A. Fischler and Robert~C. Bolles.
\newblock Random sample consensus: A paradigm for model fitting with applications to image analysis and automated cartography.
\newblock \emph{Commun. ACM}, 24\penalty0 (6):\penalty0 381–395, 1981.

\bibitem[Funes~Mora et~al.(2014{\natexlab{a}})Funes~Mora, Monay, and Odobez]{FunesMora_ETRA_2014}
Kenneth~Alberto Funes~Mora, Florent Monay, and Jean-Marc Odobez.
\newblock Eyediap: A database for the development and evaluation of gaze estimation algorithms from rgb and rgb-d cameras.
\newblock In \emph{Proc. ETRA}, 2014{\natexlab{a}}.

\bibitem[Funes~Mora et~al.(2014{\natexlab{b}})Funes~Mora, Monay, and Odobez]{eyediap_Mora_ETRA_2014}
Kenneth~Alberto Funes~Mora, Florent Monay, and Jean-Marc Odobez.
\newblock Eyediap: A database for the development and evaluation of gaze estimation algorithms from rgb and rgb-d cameras.
\newblock In \emph{Proc. ETRA}, 2014{\natexlab{b}}.

\bibitem[Ganin and Lempitsky(2015)]{ganin2015unsupervised}
Yaroslav Ganin and Victor Lempitsky.
\newblock Unsupervised domain adaptation by backpropagation.
\newblock In \emph{Proc. ICML}, 2015.

\bibitem[Gao et~al.(2022)Gao, Guo, Wang, and Zhang]{gao2022cross}
Huan Gao, Jichang Guo, Guoli Wang, and Qian Zhang.
\newblock Cross-domain correlation distillation for unsupervised domain adaptation in nighttime semantic segmentation.
\newblock In \emph{Proc. CVPT}, pages 9913--9923, 2022.

\bibitem[Guestrin and Eizenman(2006)]{model_based1}
Elias~Daniel Guestrin and Moshe Eizenman.
\newblock General theory of remote gaze estimation using the pupil center and corneal reflections.
\newblock \emph{IEEE TBE}, 53\penalty0 (6):\penalty0 1124--1133, 2006.

\bibitem[Guo et~al.(2018)Guo, Zhu, and Lei]{3ddfa_cleardusk}
Jianzhu Guo, Xiangyu Zhu, and Zhen Lei.
\newblock 3ddfa.
\newblock \url{https://github.com/cleardusk/3DDFA}, 2018.

\bibitem[Guo et~al.(2020{\natexlab{a}})Guo, Zhu, Yang, Yang, Lei, and Li]{guo2020towards}
Jianzhu Guo, Xiangyu Zhu, Yang Yang, Fan Yang, Zhen Lei, and Stan~Z Li.
\newblock Towards fast, accurate and stable 3d dense face alignment.
\newblock In \emph{Proc. ECCV}, 2020{\natexlab{a}}.

\bibitem[Guo et~al.(2020{\natexlab{b}})Guo, Yuan, Zhang, Chi, Ling, and Zhang]{dagen_guo2020}
Zidong Guo, Zejian Yuan, Chong Zhang, Wanchao Chi, Yonggen Ling, and Shenghao Zhang.
\newblock Domain adaptation gaze estimation by embedding with prediction consistency.
\newblock In \emph{Proc. ACCV}, 2020{\natexlab{b}}.

\bibitem[Hansen and Ji(2010)]{model_2_hansen2009eye}
Dan~Witzner Hansen and Qiang Ji.
\newblock In the eye of the beholder: A survey of models for eyes and gaze.
\newblock \emph{IEEE Trans. Pattern Anal. Mach. Intell.}, 32\penalty0 (3):\penalty0 478--500, 2010.

\bibitem[He et~al.(2016)He, Zhang, Ren, and Sun]{resnet_He}
Kaiming He, Xiangyu Zhang, Shaoqing Ren, and Jian Sun.
\newblock Deep residual learning for image recognition.
\newblock In \emph{Proc. CVPR}, 2016.

\bibitem[He et~al.(2020)He, Fan, Wu, Xie, and Girshick]{he2020momentum}
Kaiming He, Haoqi Fan, Yuxin Wu, Saining Xie, and Ross Girshick.
\newblock Momentum contrast for unsupervised visual representation learning.
\newblock In \emph{Proc. CVPR}, pages 9729--9738, 2020.

\bibitem[He et~al.(2019)He, Spurr, Zhang, and Hilliges]{he2019GAN_eye}
Zhe He, Adrian Spurr, Xucong Zhang, and Otmar Hilliges.
\newblock Photo-realistic monocular gaze redirection using generative adversarial networks.
\newblock In \emph{Proc. ICCV}, pages 6932--6941, 2019.

\bibitem[Holzman et~al.(1974)Holzman, Proctor, Levy, Yasillo, Meltzer, and Hurt]{holzman1974eye}
Philip~S Holzman, Leonard~R Proctor, Deborah~L Levy, Nicholas~J Yasillo, Herbert~Y Meltzer, and Stephen~W Hurt.
\newblock Eye-tracking dysfunctions in schizophrenic patients and their relatives.
\newblock \emph{Archives of general psychiatry}, 31\penalty0 (2):\penalty0 143--151, 1974.

\bibitem[Huang et~al.(2022)Huang, Guan, Xiao, Lu, and Shao]{huang2022category}
Jiaxing Huang, Dayan Guan, Aoran Xiao, Shijian Lu, and Ling Shao.
\newblock Category contrast for unsupervised domain adaptation in visual tasks.
\newblock In \emph{Proc. CVPR}, pages 1203--1214, 2022.

\bibitem[Huang et~al.(2017)Huang, Veeraraghavan, and Sabharwal]{huang2017tabletgaze}
Qiong Huang, Ashok Veeraraghavan, and Ashutosh Sabharwal.
\newblock Tabletgaze: dataset and analysis for unconstrained appearance-based gaze estimation in mobile tablets.
\newblock \emph{Machine Vision and Applications}, 28\penalty0 (5):\penalty0 445--461, 2017.

\bibitem[Jaiswal et~al.(2020)Jaiswal, Babu, Zadeh, Banerjee, and Makedon]{jaiswal2020survey}
Ashish Jaiswal, Ashwin~Ramesh Babu, Mohammad~Zaki Zadeh, Debapriya Banerjee, and Fillia Makedon.
\newblock A survey on contrastive self-supervised learning.
\newblock \emph{Technologies}, 9\penalty0 (1):\penalty0 2, 2020.

\bibitem[Jindal and Manduchi(2022)]{jindal2022gazeclr}
Swati Jindal and Roberto Manduchi.
\newblock Contrastive representation learning for gaze estimation.
\newblock \emph{Proc. NeurIPS Workshop on Gaze Meets ML}, 2022.

\bibitem[Jourabloo and Liu(2015)]{Jourabloo_2015_ICCV}
Amin Jourabloo and Xiaoming Liu.
\newblock Pose-invariant 3d face alignment.
\newblock In \emph{Proc. ICCV}, 2015.

\bibitem[Kaya et~al.(2022)Kaya, Kumar, Sarno, Ferrari, and Van~Gool]{kaya2022neural}
Berk Kaya, Suryansh Kumar, Francesco Sarno, Vittorio Ferrari, and Luc Van~Gool.
\newblock Neural radiance fields approach to deep multi-view photometric stereo.
\newblock In \emph{Proc. WACV}, pages 1965--1977, 2022.

\bibitem[Kellnhofer et~al.(2019)Kellnhofer, Recasens, Stent, Matusik, and Torralba]{gaze360_2019}
Petr Kellnhofer, Adria Recasens, Simon Stent, Wojciech Matusik, and Antonio Torralba.
\newblock Gaze360: Physically unconstrained gaze estimation in the wild.
\newblock In \emph{Proc. ICCV}, 2019.

\bibitem[Khan et~al.(2022)Khan, AlBarri, and Manzoor]{khan2022contrastive}
Adnan Khan, Sarah AlBarri, and Muhammad~Arslan Manzoor.
\newblock Contrastive self-supervised learning: a survey on different architectures.
\newblock In \emph{2022 2nd International Conference on Artificial Intelligence (ICAI)}, pages 1--6. IEEE, 2022.

\bibitem[Kingma and Ba(2015)]{adam}
Diederik~P. Kingma and Jimmy Ba.
\newblock Adam: A method for stochastic optimization.
\newblock In \emph{ICLR}, 2015.

\bibitem[Krafka et~al.(2016)Krafka, Khosla, Kellnhofer, Kannan, Bhandarkar, Matusik, and Torralba]{gazecapture}
Kyle Krafka, Aditya Khosla, Petr Kellnhofer, Harini Kannan, Suchendra Bhandarkar, Wojciech Matusik, and Antonio Torralba.
\newblock Eye tracking for everyone.
\newblock In \emph{Proc. CVPR}, 2016.

\bibitem[Lee et~al.(2022)Lee, Yun, Kim, Na, and Yoo]{lee2022latentgaze}
Isack Lee, Jun-Seok Yun, Hee~Hyeon Kim, Youngju Na, and Seok~Bong Yoo.
\newblock Latentgaze: Cross-domain gaze estimation through gaze-aware analytic latent code manipulation.
\newblock In \emph{Proc. ACCV}, pages 3379--3395, 2022.

\bibitem[Li et~al.(2017)Li, Bolkart, Black, Li, and Romero]{FLAME:SiggraphAsia2017}
Tianye Li, Timo Bolkart, Michael.~J. Black, Hao Li, and Javier Romero.
\newblock Learning a model of facial shape and expression from {4D} scans.
\newblock In \emph{Proc. SIGGRAPH Asia}, 2017.

\bibitem[Liu et~al.(2022)Liu, Bao, Xu, Wang, Liu, and Lu]{liu2022jitter}
Ruicong Liu, Yiwei Bao, Mingjie Xu, Haofei Wang, Yunfei Liu, and Feng Lu.
\newblock Jitter does matter: Adapting gaze estimation to new domains.
\newblock \emph{arXiv preprint arXiv:2210.02082}, 2022.

\bibitem[Liu et~al.(2021)Liu, Liu, Wang, and Lu]{liu2021PnP_GA}
Yunfei Liu, Ruicong Liu, Haofei Wang, and Feng Lu.
\newblock Generalizing gaze estimation with outlier-guided collaborative adaptation.
\newblock In \emph{Proc. ICCV}, 2021.

\bibitem[LLC(2022)]{agisoft_metashape}
Agisoft LLC.
\newblock Agisoft metashape.
\newblock \url{https://www.agisoft.com/}, 2022.

\bibitem[Majaranta and Bulling(2014)]{majaranta2014eye}
P{\"a}ivi Majaranta and Andreas Bulling.
\newblock Eye tracking and eye-based human--computer interaction.
\newblock \emph{Advances in physiological computing}, pages 39--65, 2014.

\bibitem[Masi et~al.(2017)Masi, Hassner, Tran, and Medioni]{7961797}
Iacopo Masi, Tal Hassner, Anh~Tuan Tran, and G{\'e}rard Medioni.
\newblock Rapid synthesis of massive face sets for improved face recognition.
\newblock In \emph{Proc. FG}, 2017.

\bibitem[Mutlu et~al.(2009)Mutlu, Shiwa, Kanda, Ishiguro, and Hagita]{mutlu2009footing}
Bilge Mutlu, Toshiyuki Shiwa, Takayuki Kanda, Hiroshi Ishiguro, and Norihiro Hagita.
\newblock Footing in human-robot conversations: how robots might shape participant roles using gaze cues.
\newblock In \emph{Proceedings of the 4th ACM/IEEE international conference on Human robot interaction}, pages 61--68, 2009.

\bibitem[Nejjar et~al.(2023)Nejjar, Wang, and Fink]{nejjar2023dare}
Ismail Nejjar, Qin Wang, and Olga Fink.
\newblock Dare-gram: Unsupervised domain adaptation regression by aligning inverse gram matrices.
\newblock \emph{arXiv preprint arXiv:2303.13325}, 2023.

\bibitem[Ohkawa et~al.(2021)Ohkawa, Yagi, Hashimoto, Ushiku, and Sato]{ohkawa_access21_style}
Takehiko Ohkawa, Takuma Yagi, Atsushi Hashimoto, Yoshitaka Ushiku, and Yoichi Sato.
\newblock Foreground-aware stylization and consensus pseudo-labeling for domain adaptation of first-person hand segmentation.
\newblock \emph{IEEE Access}, 9:\penalty0 94644--94655, 2021.

\bibitem[Ohkawa et~al.(2022)Ohkawa, Li, Fu, Furuta, Kitani, and Sato]{ohkawa_eccv22_DAhand}
Takehiko Ohkawa, Yu-Jhe Li, Qichen Fu, Ryosuke Furuta, Kris~M. Kitani, and Yoichi Sato.
\newblock Domain adaptive hand keypoint and pixel localization in the wild.
\newblock In \emph{Proc. ECCV}, 2022.

\bibitem[Park et~al.(2018)Park, Spurr, and Hilliges]{Park2018ECCV}
Seonwook Park, Adrian Spurr, and Otmar Hilliges.
\newblock Deep pictorial gaze estimation.
\newblock In \emph{Proc. ECCV}, 2018.

\bibitem[Park et~al.(2019)Park, Mello, Molchanov, Iqbal, Hilliges, and Kautz]{faze_Park2019ICCV}
Seonwook Park, Shalini~De Mello, Pavlo Molchanov, Umar Iqbal, Otmar Hilliges, and Jan Kautz.
\newblock Few-shot adaptive gaze estimation.
\newblock In \emph{Proc. ICCV}, 2019.

\bibitem[Paysan et~al.(2009)Paysan, Knothe, Amberg, Romdhani, and Vetter]{bfm09}
Pascal Paysan, Reinhard Knothe, Brian Amberg, Sami Romdhani, and Thomas Vetter.
\newblock A 3d face model for pose and illumination invariant face recognition.
\newblock In \emph{Proc. AVSS}, 2009.

\bibitem[Qin et~al.(2022)Qin, Shimoyama, and Sugano]{qin2022learning}
Jiawei Qin, Takuru Shimoyama, and Yusuke Sugano.
\newblock Learning-by-novel-view-synthesis for full-face appearance-based 3d gaze estimation.
\newblock In \emph{Proc. CVPRW}, pages 4981--4991, 2022.

\bibitem[Ravi et~al.(2020)Ravi, Reizenstein, Novotny, Gordon, Lo, Johnson, and Gkioxari]{ravi2020pytorch3d}
Nikhila Ravi, Jeremy Reizenstein, David Novotny, Taylor Gordon, Wan-Yen Lo, Justin Johnson, and Georgia Gkioxari.
\newblock Accelerating 3d deep learning with pytorch3d.
\newblock \emph{arXiv:2007.08501}, 2020.

\bibitem[Rosu and Behnke(2022)]{rosu2022neuralmvs}
Radu~Alexandru Rosu and Sven Behnke.
\newblock Neuralmvs: Bridging multi-view stereo and novel view synthesis.
\newblock In \emph{2022 International Joint Conference on Neural Networks (IJCNN)}, pages 1--7. IEEE, 2022.

\bibitem[Scalera et~al.(2021)Scalera, Seriani, Gallina, Lentini, and Gasparetto]{scalera2021human}
Lorenzo Scalera, Stefano Seriani, Paolo Gallina, Mattia Lentini, and Alessandro Gasparetto.
\newblock Human--robot interaction through eye tracking for artistic drawing.
\newblock \emph{Robotics}, 10\penalty0 (2):\penalty0 54, 2021.

\bibitem[Shrivastava et~al.(2017)Shrivastava, Pfister, Tuzel, Susskind, Wang, and Webb]{simgan17_CVPR}
Ashish Shrivastava, Tomas Pfister, Oncel Tuzel, Joshua Susskind, Wenda Wang, and Russell Webb.
\newblock Learning from simulated and unsupervised images through adversarial training.
\newblock In \emph{Proc. CVPR}, 2017.

\bibitem[Smith et~al.(2013)Smith, Yin, Feiner, and Nayar]{colombia_CAVE_0324}
Brian~A. Smith, Qi Yin, Steven~K. Feiner, and Shree~K. Nayar.
\newblock Gaze locking: Passive eye contact detection for human-object interaction.
\newblock In \emph{Proc. UIST}, 2013.

\bibitem[Sugano et~al.(2014)Sugano, Matsushita, and Sato]{learn_by_syn_Sugano_2014_CVPR}
Yusuke Sugano, Yasuyuki Matsushita, and Yoichi Sato.
\newblock Learning-by-synthesis for appearance-based 3d gaze estimation.
\newblock In \emph{Proc. CVPR}, 2014.

\bibitem[Tan et~al.(2002)Tan, Kriegman, and Ahuja]{Tan2002AppearancebasedEG}
Kar-Han Tan, David~J Kriegman, and Narendra Ahuja.
\newblock Appearance-based eye gaze estimation.
\newblock In \emph{Proc. WACV}, 2002.

\bibitem[Tarvainen and Valpola(2017)]{mean_teacher_tarvainen2017}
Antti Tarvainen and Harri Valpola.
\newblock Mean teachers are better role models: Weight-averaged consistency targets improve semi-supervised deep learning results.
\newblock \emph{Proc. NeurIPS}, 30, 2017.

\bibitem[Tuan~Tran et~al.(2017)Tuan~Tran, Hassner, Masi, and Medioni]{Tran_2017_CVPR}
Anh Tuan~Tran, Tal Hassner, Iacopo Masi, and Gerard Medioni.
\newblock Regressing robust and discriminative 3d morphable models with a very deep neural network.
\newblock In \emph{Proc. CVPR}, 2017.

\bibitem[Wang et~al.(2019{\natexlab{a}})Wang, Zhao, Su, and Ji]{wang2019generalizing}
Kang Wang, Rui Zhao, Hui Su, and Qiang Ji.
\newblock Generalizing eye tracking with bayesian adversarial learning.
\newblock In \emph{Proceedings of the IEEE/CVF conference on computer vision and pattern recognition}, pages 11907--11916, 2019{\natexlab{a}}.

\bibitem[Wang et~al.(2019{\natexlab{b}})Wang, Han, Huang, Dong, and Scott]{triplet1_wang2019multi}
Xun Wang, Xintong Han, Weilin Huang, Dengke Dong, and Matthew~R Scott.
\newblock Multi-similarity loss with general pair weighting for deep metric learning.
\newblock In \emph{Proc. CVPR}, pages 5022--5030, 2019{\natexlab{b}}.

\bibitem[Wang et~al.(2022)Wang, Jiang, Li, Ni, Dai, Li, Xiong, and Li]{wang2022contrastive}
Yaoming Wang, Yangzhou Jiang, Jin Li, Bingbing Ni, Wenrui Dai, Chenglin Li, Hongkai Xiong, and Teng Li.
\newblock Contrastive regression for domain adaptation on gaze estimation.
\newblock In \emph{Proc. CVPR}, pages 19376--19385, 2022.

\bibitem[Weidenbacher et~al.(2007)Weidenbacher, Layher, Strauss, and Neumann]{eye_2007}
Ulrich Weidenbacher, Georg Layher, P-M Strauss, and Heiko Neumann.
\newblock A comprehensive head pose and gaze database.
\newblock In \emph{3rd IET International Conference on Intelligent Environments}, pages 455--458, 2007.

\bibitem[Wood et~al.(2016{\natexlab{a}})Wood, Baltru{\v{s}}aitis, Morency, Robinson, and Bulling]{wood2016_etra}
Erroll Wood, Tadas Baltru{\v{s}}aitis, Louis-Philippe Morency, Peter Robinson, and Andreas Bulling.
\newblock Learning an appearance-based gaze estimator from one million synthesised images.
\newblock In \emph{Proc. ETRA}, 2016{\natexlab{a}}.

\bibitem[Wood et~al.(2016{\natexlab{b}})Wood, Baltrušaitis, Morency, Robinson, and Bulling]{egp.20161054}
Erroll Wood, Tadas Baltrušaitis, Louis-Philippe Morency, Peter Robinson, and Andreas Bulling.
\newblock {A 3D Morphable Model of the Eye Region}.
\newblock In \emph{EG - Posters}, 2016{\natexlab{b}}.

\bibitem[Xiao et~al.(2023)Xiao, Huang, Xuan, Ren, Liu, Guan, Saddik, Lu, and Xing]{xiao20233d}
Aoran Xiao, Jiaxing Huang, Weihao Xuan, Ruijie Ren, Kangcheng Liu, Dayan Guan, Abdulmotaleb~El Saddik, Shijian Lu, and Eric Xing.
\newblock 3d semantic segmentation in the wild: Learning generalized models for adverse-condition point clouds.
\newblock \emph{arXiv preprint arXiv:2304.00690}, 2023.

\bibitem[Xie et~al.(2020)Xie, Dai, Hovy, Luong, and Le]{xie2020unsupervised}
Qizhe Xie, Zihang Dai, Eduard Hovy, Thang Luong, and Quoc Le.
\newblock Unsupervised data augmentation for consistency training.
\newblock \emph{Advances in NIPS}, 33:\penalty0 6256--6268, 2020.

\bibitem[Xiong et~al.(2019)Xiong, Kim, and Singh]{xiong2019mixed}
Yunyang Xiong, Hyunwoo~J Kim, and Vikas Singh.
\newblock Mixed effects neural networks (menets) with applications to gaze estimation.
\newblock In \emph{Proc. CVPR}, 2019.

\bibitem[Yu and Odobez(2020)]{yu2020unsupervised}
Yu Yu and Jean-Marc Odobez.
\newblock Unsupervised representation learning for gaze estimation.
\newblock In \emph{Proc. CVPR}, 2020.

\bibitem[Yu et~al.(2019)Yu, Liu, and Odobez]{yu2019_syn_user_specific}
Yu Yu, Gang Liu, and Jean-Marc Odobez.
\newblock Improving few-shot user-specific gaze adaptation via gaze redirection synthesis.
\newblock In \emph{Proc. CVPR}, pages 11937--11946, 2019.

\bibitem[{Yuxiao Hu} et~al.(2004){Yuxiao Hu}, {Dalong Jiang}, {Shuicheng Yan}, {Lei Zhang}, and {Hongjiang zhang}]{rec_recog_2004}
{Yuxiao Hu}, {Dalong Jiang}, {Shuicheng Yan}, {Lei Zhang}, and {Hongjiang zhang}.
\newblock Automatic 3d reconstruction for face recognition.
\newblock In \emph{Proc. FG}, 2004.

\bibitem[Zhang et~al.(2022)Zhang, Liu, and Lu]{gazeonce_ZhangCVPR22}
Mingfang Zhang, Yunfei Liu, and Feng Lu.
\newblock Gazeonce: Real-time multi-person gaze estimation.
\newblock In \emph{Proc. CVPR}, pages 4197--4206, 2022.

\bibitem[Zhang et~al.(2015)Zhang, Sugano, Fritz, and Bulling]{zhang15_cvpr}
Xucong Zhang, Yusuke Sugano, Mario Fritz, and Andreas Bulling.
\newblock Appearance-based gaze estimation in the wild.
\newblock In \emph{Proc. CVPR}, 2015.

\bibitem[Zhang et~al.(2017)Zhang, Sugano, Fritz, and Bulling]{swcnn_zhang2017s}
Xucong Zhang, Yusuke Sugano, Mario Fritz, and Andreas Bulling.
\newblock It's written all over your face: Full-face appearance-based gaze estimation.
\newblock In \emph{Proc. CVPRW}, 2017.

\bibitem[Zhang et~al.(2018)Zhang, Sugano, and Bulling]{zhang18_etra}
Xucong Zhang, Yusuke Sugano, and Andreas Bulling.
\newblock Revisiting data normalization for appearance-based gaze estimation.
\newblock In \emph{Proc. ETRA}, 2018.

\bibitem[Zhang et~al.(2019)Zhang, Sugano, Fritz, and Bulling]{mpii_zhang19_pami}
Xucong Zhang, Yusuke Sugano, Mario Fritz, and Andreas Bulling.
\newblock Mpiigaze: Real-world dataset and deep appearance-based gaze estimation.
\newblock \emph{IEEE Trans. Pattern Anal. Mach. Intell.}, 41\penalty0 (1):\penalty0 162--175, 2019.

\bibitem[Zhang et~al.(2020{\natexlab{a}})Zhang, Park, Beeler, Bradley, Tang, and Hilliges]{Zhang2020ETHXGaze}
Xucong Zhang, Seonwook Park, Thabo Beeler, Derek Bradley, Siyu Tang, and Otmar Hilliges.
\newblock Eth-xgaze: A large scale dataset for gaze estimation under extreme head pose and gaze variation.
\newblock In \emph{Proc. ECCV}, 2020{\natexlab{a}}.

\bibitem[Zhang et~al.(2020{\natexlab{b}})Zhang, Sugano, Bulling, and Hilliges]{region_selection_Zhang2020}
Xucong Zhang, Yusuke Sugano, Andreas Bulling, and Otmar Hilliges.
\newblock Learning-based region selection for end-to-end gaze estimation.
\newblock In \emph{Proc. BMVC}, 2020{\natexlab{b}}.

\bibitem[Zheng et~al.(2020)Zheng, Park, Zhang, Mello, and Hilliges]{sted_Zheng2020NeurIPS}
Yufeng Zheng, Seonwook Park, Xucong Zhang, Shalini~De Mello, and Otmar Hilliges.
\newblock Self-learning transformations for improving gaze and head redirection.
\newblock In \emph{Proc. NeurIPS}, 2020.

\bibitem[Zhou et~al.(2017)Zhou, Lapedriza, Khosla, Oliva, and Torralba]{zhou2017places}
Bolei Zhou, Agata Lapedriza, Aditya Khosla, Aude Oliva, and Antonio Torralba.
\newblock Places: A 10 million image database for scene recognition.
\newblock \emph{IEEE Trans. Pattern Anal. Mach. Intell.}, 2017.

\bibitem[Zhou et~al.(2020)Zhou, Liu, Liu, Liu, and Wang]{zhou2020rotate}
Hang Zhou, Jihao Liu, Ziwei Liu, Yu Liu, and Xiaogang Wang.
\newblock Rotate-and-render: Unsupervised photorealistic face rotation from single-view images.
\newblock In \emph{Proc. CVPR}, 2020.

\bibitem[Zhu et~al.(2016)Zhu, Lei, Liu, Shi, and Li]{Zhu_2016_CVPR}
Xiangyu Zhu, Zhen Lei, Xiaoming Liu, Hailin Shi, and Stan~Z. Li.
\newblock Face alignment across large poses: A 3d solution.
\newblock In \emph{Proc. CVPR}, 2016.

\bibitem[Zhu et~al.(2019)Zhu, Liu, Lei, and Li]{zhu2017face}
Xiangyu Zhu, Xiaoming Liu, Zhen Lei, and Stan~Z Li.
\newblock Face alignment in full pose range: A 3d total solution.
\newblock \emph{IEEE Trans. Pattern Anal. Mach. Intell.}, 41\penalty0 (1):\penalty0 78--92, 2019.

\bibitem[Zollh{\"o}fer et~al.(2018)Zollh{\"o}fer, Thies, Garrido, Bradley, Beeler, P{\'e}rez, Stamminger, Nie{\ss}ner, and Theobalt]{Zollhfer2018StateOT}
Michael Zollh{\"o}fer, Justus Thies, Pablo Garrido, Derek Bradley, Thabo Beeler, Patrick P{\'e}rez, Marc Stamminger, Matthias Nie{\ss}ner, and Christian Theobalt.
\newblock State of the art on monocular 3d face reconstruction, tracking, and applications.
\newblock \emph{Computer Graphics Forum}, 37\penalty0 (2):\penalty0 523--550, 2018.

\end{thebibliography}
}

\end{document}